\documentclass[acmsmall,screen]{acmart}
\usepackage{bm}
\AtBeginDocument{%
  }

\setcopyright{acmlicensed}
\copyrightyear{2018}
\acmYear{2018}
\acmDOI{XXXXXXX.XXXXXXX}

\acmJournal{JACM}
\acmVolume{37}
\acmNumber{4}
\acmArticle{111}
\acmMonth{8}

\begin{document}

\title{UrbanMind: Towards Urban General Intelligence via Tool-Enhanced Retrieval-Augmented Generation and Multilevel Optimization}

\author{Kai Yang}
\email{kaiyang@tongji.edu.cn}
\affiliation{%
  \institution{Tongji University}
  \state{Shanghai}
  \country{China}
}
\author{Zelin Zhu}
\affiliation{%
  \institution{Tongji University}
  \state{Shanghai}
  \country{China}
}
\author{Chengtao Jian}
\affiliation{%
  \institution{Tongji University}
  \state{Shanghai}
  \country{China}
}
\author{Hui Ma}
\affiliation{%
  \institution{Xinjiang University}
  \state{Xinjiang}
  \country{China}
}
\author{Shengjie Zhao}
\affiliation{%
  \institution{Tongji University}
  \state{Shanghai}
  \country{China}
}

\author{Xiaozhou Ye}
\affiliation{%
  \institution{Asiainfo Technologies}
  \state{Beijing}
  \country{China}
}
\author{Ye Ouyang}
\affiliation{%
  \institution{Asiainfo Technologies}
  \state{Beijing}
  \country{China}
}


\begin{abstract}
Urban general intelligence (UGI) refers to the capacity of AI systems to autonomously perceive, reason, and act within dynamic and complex urban environments. In this paper, we introduce UrbanMind, a tool-enhanced retrieval-augmented generation (RAG) framework designed to facilitate UGI. Central to UrbanMind is a novel architecture based on Continual Retrieval-Augmented MoE-based LLM (C-RAG-LLM), which dynamically incorporates domain-specific knowledge and evolving urban data to support long-term adaptability. The architecture of C-RAG-LLM aligns naturally with a multilevel optimization framework, where different layers are treated as interdependent sub-problems. Each layer has distinct objectives and can be optimized either independently or jointly through a hierarchical learning process. The framework is highly flexible, supporting both end-to-end training and partial layer-wise optimization based on resource or deployment constraints. To remain adaptive under data drift, it is further integrated with an incremental corpus updating mechanism. Evaluations on real-world urban tasks of a variety of complexity verify the effectiveness of the proposed framework. This work presents a promising step toward the realization of general-purpose LLM agents in future urban environments.
\end{abstract}


\maketitle

\section{Introduction and Background}
\subsection{Motivation for Urban General Intelligence}

The rapid expansion of urbanization presents not only new opportunities but also significant challenges for modern cities. Urban environments are inherently dynamic and complex, characterized by the continuous evolution of infrastructures and the frequent occurrence of unpredictable events\cite{bettencourt2021introduction}. Traditional AI systems\cite{nama2021machine,ramana2023vision}, which are often developed for static tasks, exhibit fundamental limitations when applied to such non-stationary settings. To achieve Artificial General Intelligence (AGI), systems must be dynamic. Unlike narrow AI, AGI requires strong generalization and the ability to adapt across different tasks\cite{goertzel2014artificial}. Recent studies highlight that multimodal understanding and the continuous adaption are key to building such systems\cite{fei2022towards,li2025survey}. In addition, for areas like IoT AGI must process real-time data and make decisions accordingly\cite{latif2023artificial,dou2023towards,li2024artificial}. Achieving UGI can significantly improve the performance of critical urban tasks, including traffic management\cite{ravish2021intelligent}, public safety\cite{mahor2023iot}, disaster response\cite{khatoon2022social}.

Realizing UGI requires fundamental advancements in urban foundation model\cite{zhang2024urban}, continual learning\cite{yang2024continual}, dynamic knowledge integration\cite{fadhel2024comprehensive}, and context-aware decision-making\cite{pacheco2022systematic}. Such capabilities are essential to support the long-term evolution of urban systems toward safer and smarter environments. Therefore, UGI represents not only a technical advancement but also a critical step in redefining the role of AI within the fabric of future cities.

Despite its transformative potential, realizing UGI imposes significant challenges. Urban environments are characterized by non-stationary data distributions that evolve due to factors such as seasonal variations\cite{liu2019seasonal} and infrastructure developments\cite{abbes2016urban}. Designing AI systems that can adapt to such changes without catastrophic forgetting remains a major obstacle\cite{yang2024continual}. Additionally, urban data is inherently heterogeneous and noisy, further complicating reliable knowledge extraction, reasoning, and decision-making processes\cite{yang2022survey}. Conventional machine learning paradigms, which assume static training and deployment conditions, are not well suited for the continuous and adaptive nature of urban environments.

In this paper, we propose a framework called UrbanMind, which leverages a multilevel optimization paradigm to jointly address the core requirements essential for realizing UGI. 

\subsection{RAG and Continual Learning}

RAG\cite{lewis2020retrieval} has emerged as an effective meanings for enhancing the reasoning and generation capabilities of LLMs by integrating external knowledge sources. Unlike traditional models that rely solely on internal parameters to store factual knowledge\cite{floridi2020gpt,touvron2023llama}, RAG dynamically retrieves relevant information from external corpora to assist in generation. This mechanism allows the model to remain lightweight while maintaining access to a continually expanding and domain-specific knowledge base. Recent studies have demonstrated the benefits of RAG in enhancing factual accuracy and adaptability across a wide range of  tasks\cite{xia2024rule,ding2024realgen,xu2024retrieval}. However, most existing RAG systems are designed for static retrieval settings and do not directly address the challenges posed by non-stationary environments or continual updates to the knowledge source.

RAG also enhances multi-agent collaboration by enabling agents to access external knowledge in real time. For example, LLMs agents often rely on RAG to retrieve relevant facts from knowledge graphs or databases. A recent example is CLADD\cite{lee2025rag}, RAG enables specialized agents, such as those analyzing molecular structures or querying knowledge graphs to retrieve and share domain specific information, allowing the system to generate more accurate and context-aware answers. Besides, RAG bridges LLMs with tools as retrieved content can guide tool use, and tool outputs can be reintegrated into the LLM’s context.

Continual learning\cite{wang2024comprehensive} aims to develop models that can incrementally learn from new data streams while preserving knowledge acquired from previous experiences. It addresses the critical limitation of traditional machine learning paradigms, where retraining from scratch or fine-tuning on new data often leads to catastrophic forgetting\cite{nguyen2019toward}. Various strategies, such as regularization\cite{pomponi2020efficient}, memory replay\cite{wang2022memory}, and dynamic architectural expansion\cite{gao2022efficient}, have been proposed to mitigate forgetting and support stable learning over time. While continual learning has shown promise in areas such as robotics\cite{lesort2020continual}, natural language processing\cite{biesialska2020continual} and vision\cite{qu2025recent}, most approaches assume access to well-structured task boundaries and stable data flows. The integration of continual learning with RAG, particularly under dynamic and evolving data distributions typical of urban systems, remains an underexplored area.

Updating the knowledge base plays an essential role in maintaining the performance and accuracy of RAG systems. It involves two key components: update triggering and the implementation of update strategies. Update triggering is driven by factors such as timeliness, user feedback, and system performance metrics. In domains with rapidly evolving information, such as finance and meteorology, continuous analysis of user queries and feedback can reveal knowledge gaps—particularly when there is a high frequency of queries yielding inadequate or irrelevant responses. Additionally, monitoring system-level indicators\cite{roychowdhury2024evaluation},including retrieval recall, precision, and response satisfaction, can help detect knowledge degradation. A noticeable decline in these metrics often signals that the underlying knowledge base is outdated or incomplete, necessitating updates. In response, various updating strategies can be adopted. Automated updates are particularly suitable for structured data sources with regular update patterns and constrained memory, with mechanisms such as sliding windows enabling dynamic memory maintenance\cite{fan2025research}. Manual updates remain crucial in scenarios requiring domain expertise or professional validation. Moreover, machine learning-assisted methods offer scalable solutions by analyzing incoming data streams, classifying content, and identifying novel knowledge elements\cite{yang2019mares}. For example, DR-RAG\cite{hei2024dr} introduces a two-stage retrieval mechanism that adaptively selects relevant documents based on user queries, providing an effective approach for maintaining contextual relevance during knowledge base updates.

\subsection{Urban Foundational Model}
Urban foundational models are large-scale pre-trained models designed to capture the broad distributions and dynamics inherent in urban environments\cite{zhang2024urban}. Similar to general-purpose foundation models in natural language processing and vision, urban foundational models are trained on diverse multimodal datasets encompassing transportation patterns, public safety reports, environmental sensor data, land use information, and social behavioral traces\cite{balsebre2023cityfm,zhang2024urban}. The objective is to learn generalizable representations that can support a wide range of downstream urban tasks with minimal task-specific fine-tuning. By pre-training across multiple domains and modalities, these models serve as universal foundations for reasoning and decision-making in complex urban systems.

Training urban foundational models presents several significant challenges due to the complexity and heterogeneity of urban data\cite{longley2004spatial}. First, the multimodal nature of urban information ranging from structured spatial data to unstructured textual reports requires the development of unified encoding architectures capable of fusing diverse data types effectively. Second, urban datasets often suffer from noise and missing values which can impair the quality of learned representations and limit generalization\cite{zhang2024urban}. In addition, achieving scalability while maintaining fine-grained temporal and spatial resolution is computationally intensive, necessitating efficient data management and training strategies\cite{bibri2021data}. Finally, ensuring that pre-trained models remain adaptable to continual retrieval and evolving urban contexts introduces additional demands on model regularization and dynamic fine-tuning capabilities\cite{freire2021artificial}. Addressing these challenges plays an essential role in constructing reliable urban foundational models.

\subsection{Tool-Enhanced Retrieval-Augmented Generation}

To bridge the gap between foundation models and domain-specific knowledge or real-world actions, recent research has focused on integrating LLMs with external tools to form LLM-empowered agents\cite{zhao2024let}. This paradigm endows LLM agents with enhanced capabilities beyond their intrinsic parameters, offering  a variety of benefits. 

Among various tool calling strategies, Tool-Enhanced Retrieval Augmented Generation represents a foundational implementation wherein the LLM formulates a query, retrieves relevant documents, and incorporates the retrieved evidence into its generation process\cite{huang2024tool, li2025review}. Recent advances further explore integrating RAG with optimizing tool calling\cite{wu2024avatar,patil2024gorilla},where LLMs call tools more wisely or efficiently and LLMs generate API-compatible queries for search engines or domain tools\cite{huang2024tool}, demonstrating superior performance in knowledge-intensive domains.

In urban environments, RAG-based tool calling can provide an effective means for enabling LLM-agent architectures to address complex and multimodal problems. Built on top of Urban Foundation Models and Tool-Enhanced Retrieval Augmented Generation, such agents can leverage a diverse set of external tools, including for example traffic simulators, spatio-temporal databases, weather forecasting modules, remote sensing APIs, to perceive, reason, and act within the urban ecosystem. 

\subsection{Related Work}
AGI  refers to synthetic intelligence with broad scope and strong generalization capabilities, fundamentally different from narrow AI with limited adaptability\cite{goertzel2014artificial}. Recent research has begun exploring AGI applications across various domains. In education, AGI could enable personalized and adaptive learning experiences\cite{latif2023artificial}; in IoT, AGI is expected to support real-time, context-aware decision-making beyond current narrow solutions\cite{dou2023towards}. Key capabilities of AGI include multimodal understanding, interactivity, and personalization are essential for advancing toward more adaptive and human-aligned AI systems\cite{fei2022towards,li2025survey,morris2024position}.

The application of AGI\cite{xu2023urban} concepts to urban systems is still at an early stage of exploration. Existing efforts have largely focused on building specialized AI models for distinct urban tasks, such as traffic management\cite{ait2022fusion} and public safety surveillance\cite{srivastava2017safety}. While these models have demonstrated strong task-specific performance, they lack the flexibility and cross-domain reasoning capabilities necessary for true general intelligence. Recent advances in LLMs, reinforcement learning\cite{wang2024reinforcement}, and multi-agent systems\cite{han2024llm} have opened up new avenues for broader urban decision-making. However, the majority of current approaches operate under static assumptions and do not address the challenges imposed by dynamic and evolving urban environments.

Early attempts at integrating continual learning into urban applications have primarily focused on incremental model retraining without systematic mechanisms for long-term knowledge preservation or cross-domain reasoning\cite{chen2021trafficstream,rui2023dilrs}. Furthermore, the potential of RAG frameworks to enhance continual learning in urban contexts has not been thoroughly investigated. These limitations motivate the need for new architectures that combine retrieval-based knowledge integration with continual adaptation.

Multilevel optimization has recently attracted increasing attention in machine learning due to its ability to model nested decision processes encountered in applications such as meta-learning, hyperparameter tuning\cite{kim2025stochastic} and hierarchical reinforcement learning\cite{gammelli2023graph}. Classical bilevel optimization methods, which optimize an outer objective subject to the solution of an lower-level problem, have been widely studied and form the basis for many of these developments\cite{nisha2022bilevel,chen2024robust}. However, most existing work focuses on settings where task distributions are static and data availability is assumed to be complete, making the resulting algorithms unsuitable for dynamic environments like urban systems.

To address evolving data distributions and structural shifts, a few recent studies\cite{ali2024tri,duran2025adaptive} have begun exploring extensions of multilevel optimization to continual and adaptive settings. Nevertheless, current methods often either assume access to all task information simultaneously or rely on rigid update schedules that limit their responsiveness to rapid changes. Moreover, the integration of external knowledge retrieval within multilevel optimization frameworks remains largely unexplored. These gaps motivate the need for new formulations that can jointly manage retrieval, continual learning, and model adaptation in a unified multilevel structure.

\subsection{Summary of Contributions}

This paper proposes a novel framework, UrbanMind, for advancing UGI by integrating retrieval-based knowledge acquisition with continual learning under a multilevel optimization perspective. The key contributions are summarized as follows.

\begin{itemize}
\item Tool-Enhanced RAG with Continual Learning: We propose UrbanMind, a tool-enhanced RAG framework tailored for UGI. UrbanMind implements a C-RAG-LLM architecture that integrates continual learning and tool-augmented reasoning to support dynamic, context-aware decision-making in complex urban environments. The system can continuously retrieve domain-specific knowledge and incrementally adapts to evolving data distributions. Moreover, this framework can be deployed in a cloud-edge distributed manner, supporting efficient computation, real-time responsiveness, and privacy preservation by processing sensitive data locally on edge devices.
\item Multilevel Optimization with Expert Modularity: We introduce a novel multilevel optimization formulation for UGI. To the best of our knowledge, such a framework has not been previously explored in this context. This formulation provides a unified perspective that jointly models continual retrieval, knowledge integration, and model adaptation, and is closely aligned with the Mixture-of-Experts (MoE) architecture, where expert modules are selectively optimized at different levels. The proposed approach enables principled coordination across components, ensuring stable learning under non-stationary and Out-Of-Distribution (OOD) data distributions in evolving urban environments. Notably, this multilevel optimization framework is highly flexible, supporting either end-to-end optimization or selective tuning of specific components based on available computational resources and deployment requirements.
\end{itemize}

We also implement the proposed UrbanMind and conduct evaluations on real-world urban tasks, demonstrating that our proposed framework achieves superior performance compared to baseline approaches.

\section{UrbanMind for Urban General Intelligence}
\begin{figure}[h]
  \centering
  \includegraphics[width=0.8\linewidth,
                    trim=0cm 1cm 4cm 0cm,
                    clip                 
                    ]{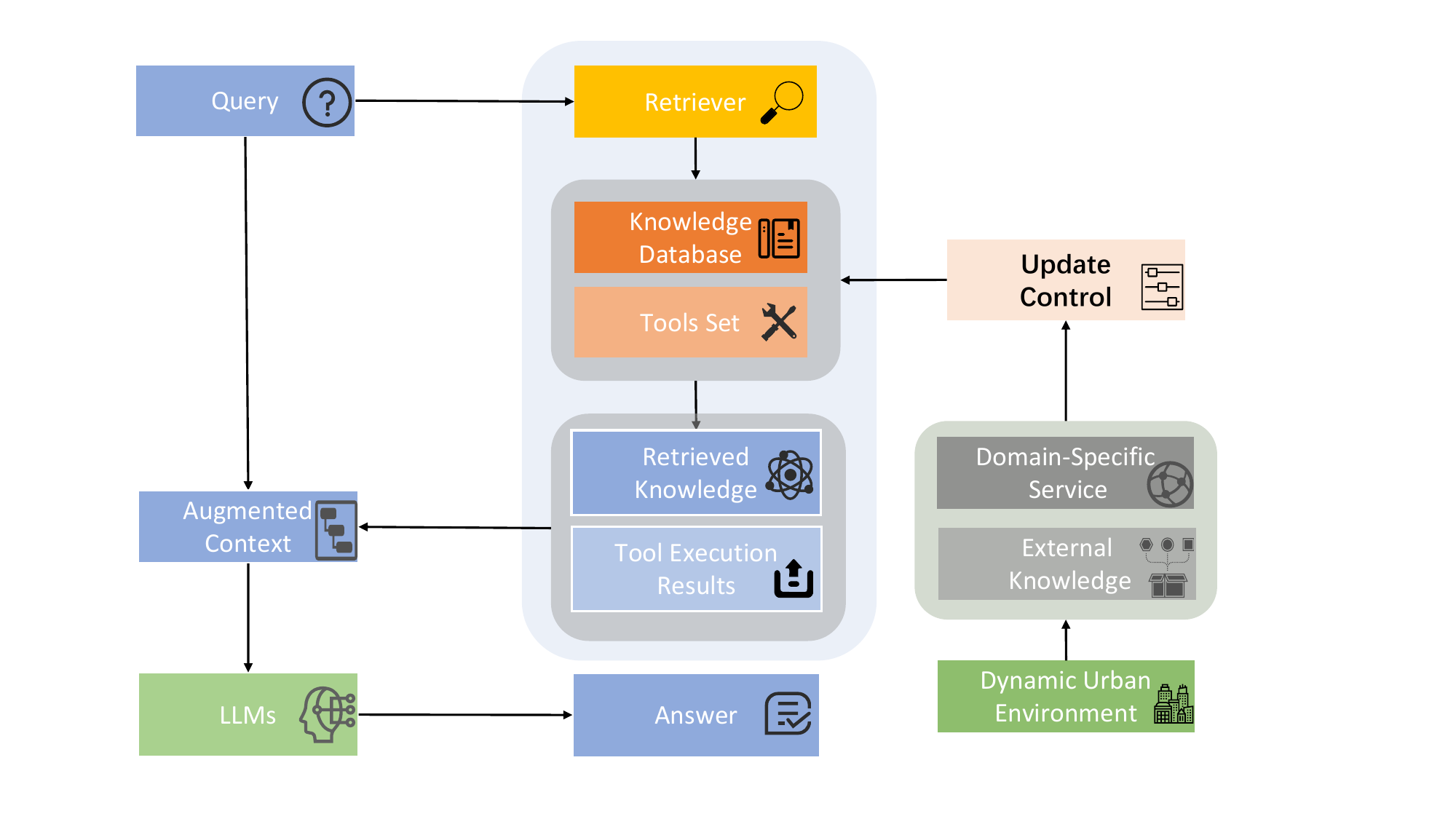}
  \caption{Tool-Enhanced RAG with Continual Learning}
  \Description{continual rag overview}
\end{figure}

\subsection{Problem Definition}

We consider a dynamic urban environment where the data distribution evolves over time due to external factors such as infrastructure changes, policy shifts, and societal behaviors. Let $\mathcal{X}_t$ denote the input space and $\mathcal{Y}_t$ the corresponding output space at time step $t$, respectively. At each time step, the AI system receives a data stream $\mathcal{D}_t = \{(x_t^i, y_t^i)\}_{i=1}^{n_t}$, where $n_t$ denotes the number of samples collected during period $t$. Unlike classical supervised learning, where the data distribution is assumed to be stationary, here the distribution $\mathcal{P}_t(x, y)$ underlying $\mathcal{D}_t$ is non-stationary, and both the input characteristics and the output semantics may change over time. The goal is to learn a predictive function $f_{\theta_t}$ parameterized by $\theta_t$, which maintains high performance across all past and present distributions without retraining from scratch.

To enable continual adaptation, we incorporate a retrieval-augmented mechanism into the learning process. Specifically, at each time step t, given a query $x_t$, the agent retrieves a set of external knowledge entries $\mathcal{R}_t(x_t) =\{r_t^j\}_{j=1}^{k}$ from a dynamic knowledge base $\mathcal{K}_t$, where k denotes the number of items retrieved. The information retrieved is used to augment the model input or intermediate representations, allowing the predictive function to be conditioned not only on the raw input $x_t$ but also on relevant contextual knowledge. Formally, the predictive function is expressed as $f_{\theta_t}(x_t, \mathcal{R}_t(x_t))$, and the learning objective at each time step is to minimize the expected loss $\mathbb{E}_{(x,y) \sim \mathcal{P}_t}[\ell(f_{\theta_t}(x, \mathcal{R}_t(x)), y)]$, where $\ell(\cdot)$ denotes a task-specific loss function.

The continual learning objective requires that the model parameters $\theta_t$ evolve across time to accommodate new tasks while preserving performance on previously seen tasks. To this end, the training process is formulated as a multilevel optimization problem. The first level optimizes retrieval mechanisms $\mathcal{R}_t(\cdot)$, the second level optimizes the model adaptation $f_{\theta_t}$ based on retrieved knowledge, and the third level coordinates knowledge database updating and forward knowledge transfer across time steps. The formal problem can be stated as finding a sequence $\{\theta_t\}_{t=1}^T$ and retrieval policies $\{\mathcal{R}_t\}_{t=1}^T$ that jointly minimize cumulative loss across all time steps, subject to stability constraints that prevent catastrophic forgetting and ensure continual improvement.

The proposed UrbanMind framework consists of three main components: (i) continual knowledge integration module, (ii) dynamic retrieval module, and (iii) adaptive model updating module. These components are designed to jointly optimize the retrieval, integration, and adaptation processes over evolving data distributions, thereby enabling long-term stability, forward transfer, and robust decision-making under non-stationary conditions. Each component operates within a multilevel optimization hierarchy to ensure coordinated and efficient learning.

At each time step $t$, the dynamic retrieval module is responsible for identifying and extracting relevant information $\mathcal{R}_t(x_t)$ from the evolving knowledge base $\mathcal{K}_t$, based on the input query $x_t$. The retrieval process is adaptive, allowing the system to dynamically incorporate the most relevant domain-specific knowledge. The continual knowledge integration module then fuses the retrieved information with the original query, producing an augmented context that serves as the basis for subsequent prediction or decision-making.

The adaptive model updating module incrementally refines the model parameters $\theta_t$ at given time instance $t$ to incorporate new information while preserving critical capabilities acquired from previous tasks. This is achieved through a multilevel optimization strategy, where the retrieval module and model update module are optimized jointly. The overall framework aims to minimize cumulative predictive loss while enforcing stability constraints that may mitigate catastrophic forgetting. By systematically coordinating retrieval, integration, and adaptation, the UrbanMind framework provides a resilient foundation for achieving UGI.

One fundamental challenge in continual retrieval lies in maintaining retrieval relevance and consistency over time. As the knowledge base $\mathcal{K}_t$ evolves over time, retrieval strategies that are static or trained on historical distributions may rapidly degrade in effectiveness. It becomes necessary to design retrieval mechanisms that not only adapt to the dynamic structure of $\mathcal{K}_t$ but also preserve semantic consistency with previously retrieved knowledge. 

Please note that, in the proposed framework, retriever strategy optimization, knowledge database updating, and model adaptation are decoupled and operated across different time scales to balance responsiveness, stability, and computational efficiency. Retriever strategy optimization is executed at a relatively short time scale, frequently adjusting retrieval policies based on immediate task relevance and feedback from model performance. This enables the system to maintain high retrieval precision as the query distribution evolves. In contrast, knowledge updating mechanisms may operate at an intermediate time scale, periodically incorporating new data into the knowledge base while validating and pruning outdated or low-relevance entries. This ensures that the retrieval corpus remains current without introducing instability from overly frequent modifications. Model adaptation usually occurs at the longest time scale, where fine-tuning is applied to avoid overfitting to transient data shifts and to mitigate catastrophic forgetting. This multi-timescale design allows the framework to adapt dynamically to new information while preserving long-term learning stability and computational scalability.
However, while the proposed framework generally adheres to the described multi-timescale paradigm, where retrieval strategy optimization operates most frequently, followed by knowledge base updating and then model adaptation. Such a hierarchy may invert in certain application scenarios due to domain-specific requirements. For instance, in traffic-prediction \cite{ma2023cellular,ma2023metastnet,dou2024anomaly}, where traffic conditions vary rapidly, the knowledge base must be updated almost in real time to incorporate the latest traffic indicators. In such cases, knowledge updating operates at the shortest time scale to ensure that the retrieval process accesses the most current information, even more frequently than retrieval strategy optimization. Conversely, in highly dynamic dialogue systems for personalized education, the user's interaction patterns and feedback may rapidly evolve \cite{zhang2024cppo}. Here, model adaptation may occur on a shorter timescale than knowledge updates or retrieval adjustments, especially when personalized fine-tuning is necessary to ensure responsiveness and effectiveness. These examples highlight that, in practice, the temporal scheduling of updates must be flexibly adapted to the characteristics of specific tasks and domains.

\subsection{Background: Naive RAG Pipeline}

RAG is a widely adopted framework for enhancing the reasoning capabilities of large language models (LLMs) by integrating external knowledge sources \cite{lewis2020retrieval}. Before delving into the proposed UrbanMind framework, we first introduce the Naive RAG pipeline, which serves as a baseline for understanding the RAG and its limitations in dynamic urban environments.

\begin{figure}[h]
\centering
\includegraphics[width=\linewidth,
                    trim=0cm 4cm 0cm 3cm,
                    clip                 
]{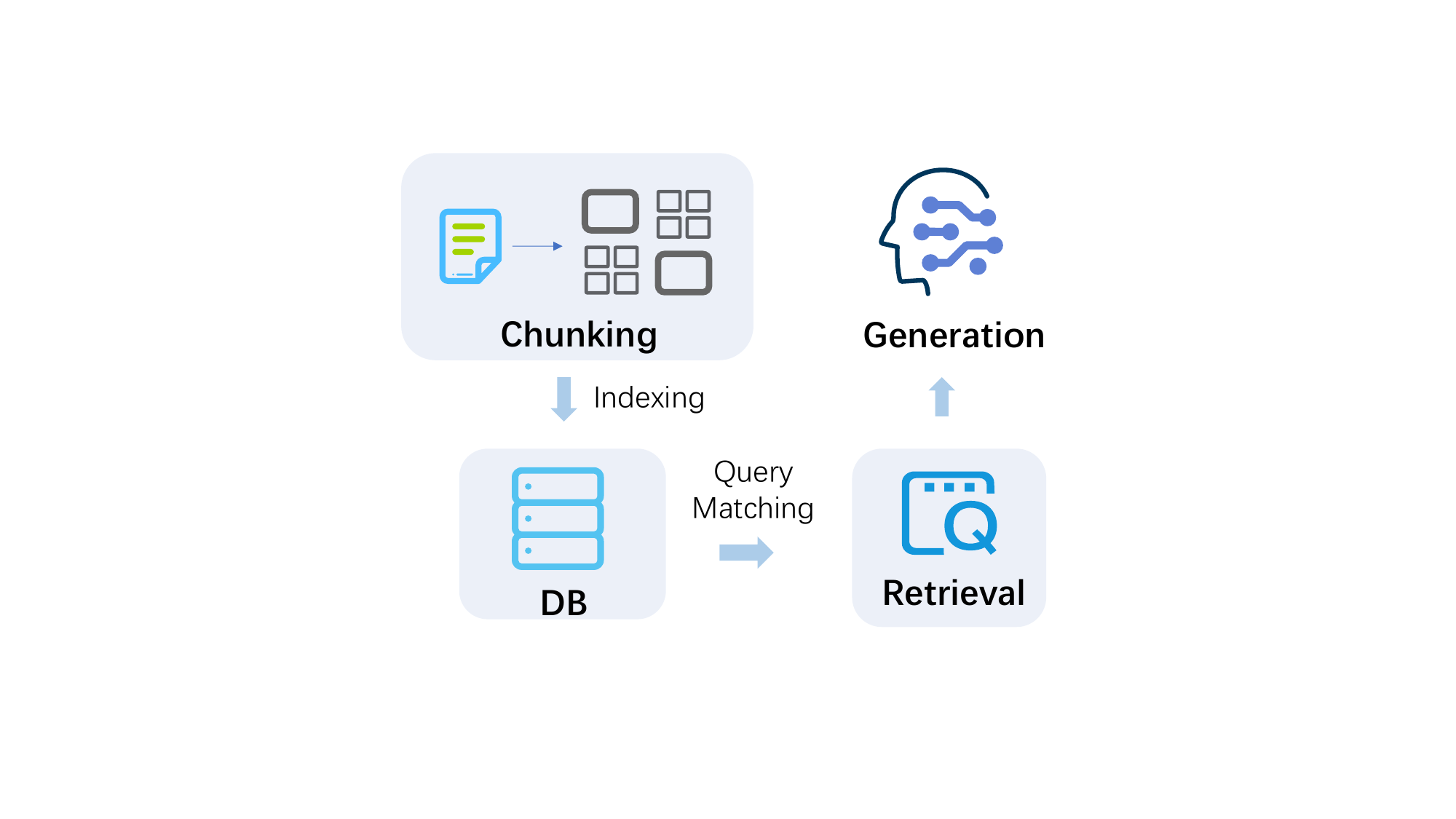}
\caption{Naive RAG Pipeline Workflow}
\Description{A diagram illustrating the Naive RAG pipeline, including chunking, database storage, retrieval, and generation stages.}
\label{fig:naive_rag_pipeline}
\end{figure} 

The Naive RAG pipeline, as depicted in Figure \ref{fig:naive_rag_pipeline}, consists of four key stages designed to incorporate external knowledge into the generation process. The workflow begins with the chunking phase, where raw documents, such as textual reports or structured datasets are segmented into smaller, semantically coherent chunks. This segmentation ensures that the knowledge is broken down into manageable units suitable for efficient storage and retrieval. Next, these chunks are indexed and stored in a database, forming a static knowledge repository that can be queried later.

In the retrieval phase, a user query, e.g., a question about urban traffic conditions is encoded into a vector representation using a pre-trained encoder, such as BERT \cite{devlin2019bert}. The encoded query is then used to search the vector database, retrieving the top-K most relevant chunks based on similarity metrics, typically cosine similarity between the query and chunk embeddings. These retrieved chunks provide external context that is critical for grounding the model’s reasoning in factual knowledge. Finally, in the generation phase, the retrieved chunks are combined with the original query and fed into a large language model, which generates a response by leveraging both the query and the retrieved knowledge. 

\subsection{Urban Intelligence Tasks}

Urban intelligence tasks encompass a broad spectrum of applications that demand AI systems capable of reasoning and operating effectively in dynamic environments. As discussed in \cite{xu2023urban}, the UGI foundation platform has been applied to various urban domains, including transportation and urban economy. Representative tasks include conducting travel surveys within transportation systems \cite{elizalde2019travel}, selecting optimal business sites in business intelligence \cite{liu2023knowsite}, formulating policies in urban economic systems \cite{kourtit2021city}, and managing emergencies in urban society \cite{golpayegani2021urban}. These tasks can broadly categorized into three major domains, i.e., \textit{public safety management}, \textit{transportation systems}, and \textit{urban planning and development}. Each domain presents unique data characteristics, operational constraints, and decision-making requirements that influence the design and deployment of continual learning and retrieval-augmented frameworks.

\textit{Public safety management} encompasses tasks such as threat detection, emergency response coordination, and predictive risk assessment for urban populations. Practical examples include the timely identification of infectious disease outbreaks through hospital reports or social media analysis, and early flood warnings enabled by monitoring river water levels using hydrological sensors. Similarly, anomalous patterns in air quality or radiation levels may indicate emerging environmental hazards. These tasks rely on heterogeneous data sources, including surveillance feeds, incident reports, social media streams, and environmental sensing infrastructures \cite{fortin2021use,poredi2023enhance}.

\textit{Transportation intelligence} focus primarily on tasks such as optimizing traffic flow and predicting congestion patterns. These tasks are often characterized by real-time data streams generated from heterogeneous sources such as sensors, GPS devices, and traffic cameras \cite{nama2021machine,ravish2021intelligent}. In broader applications, transportation intelligence also encompasses low-altitude logistics e.g., drone-based delivery, railway logistics, and highway freight systems. The underlying data distributions are subject to rapid fluctuations driven by daily commuting patterns, weather conditions, and special events, necessitating models that can quickly adapt without losing historical knowledge of baseline traffic behaviors.

\textit{Urban planning and development}, in contrast, typically operate on longer time scales and involve the integration of census data, land use maps, and environmental assessments \cite{chen2021intelligent}. These tasks require AI systems to reason over structured, semi-structured, and unstructured data formats. Collectively, the diversity across these categories imposes stringent requirements on retrieval accuracy, continual learning stability, and adaptive reasoning capabilities.

While these domains differ in timescales and data modalities, they collectively highlight a common challenge: urban data is inherently dynamic and non-stationary. Temporal variations arise from periodic patterns (e.g., commuting cycles), sudden disruptions (e.g., emergencies), and gradual changes (e.g., urbanization). Spatial heterogeneity stems from differences in geography, infrastructure, and localized behavior. Meanwhile, contextual shifts reflect the evolving nature of societal, environmental, and policy factors. These characteristics result in non-stationary data streams that challenge static learning paradigms.

To address these challenges, urban intelligence systems must support continual knowledge integration and adaptive retrieval. Retrieval mechanisms must dynamically adjust to the evolving knowledge base, ensuring relevance and robustness against semantic drift. Integration pipelines must handle noisy, incomplete, and potentially conflicting signals while maintaining alignment with historical knowledge. In addition, these processes must operate under strict latency requirements and resource constraints to enable real-time urban decision-making. The ability to continually adapt while preserving accumulated knowledge is thus fundamental to sustaining high-level reasoning in complex, real-world urban environments.

\subsection{Framework Overview}
\begin{figure}[h]
  \centering
  \includegraphics[width=0.8\linewidth,
                    trim=0cm 6cm 0cm 3cm,
                    clip                 
        ]{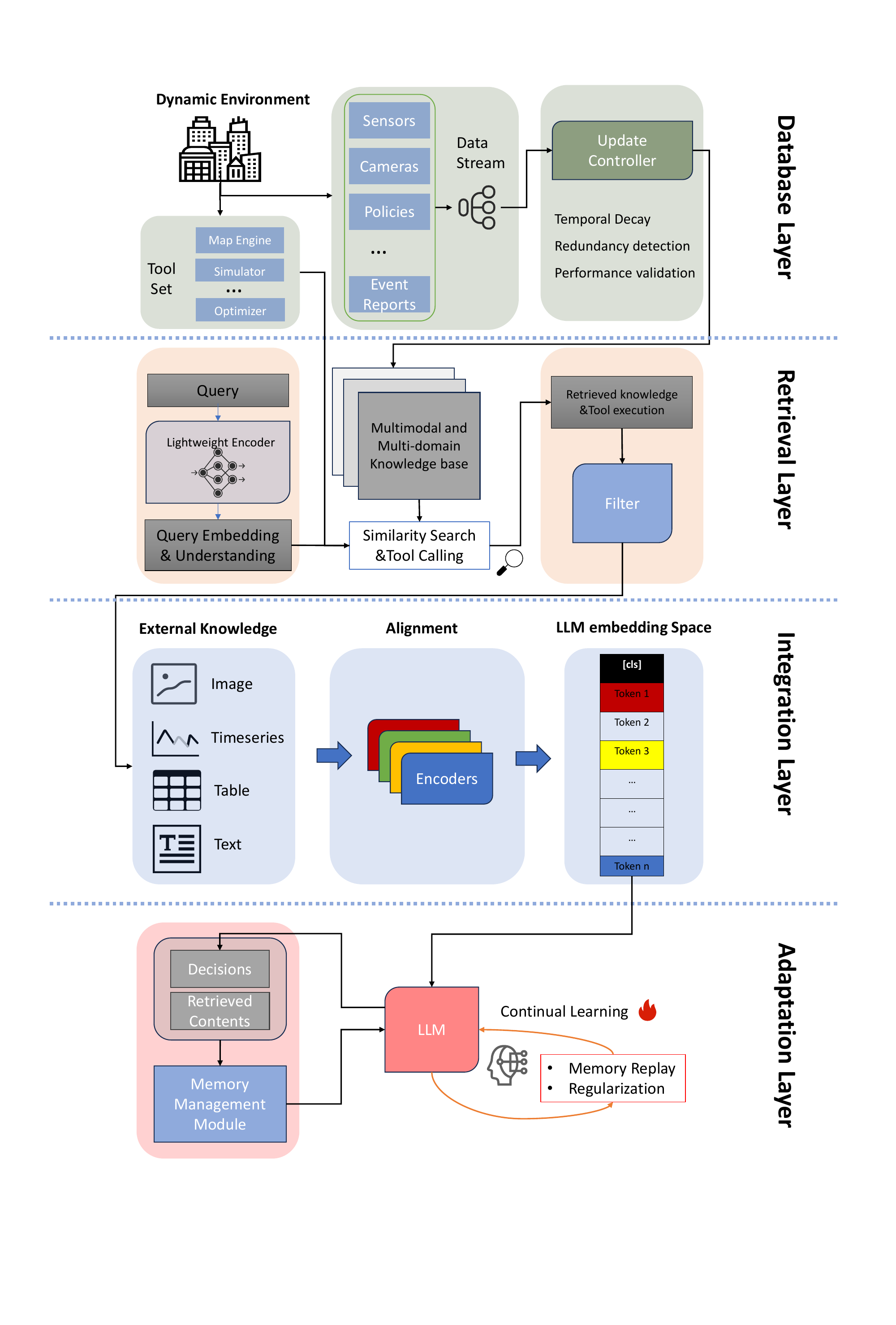}
  \caption{UrbanMind Framework}
  \Description{continual rag overview}
\end{figure}
The proposed UrbanMind framework is designed to address the challenges of dynamic knowledge acquisition, continual adaptation, and robust decision-making in non-stationary urban environments. The system architecture is organized into four interconnected layers, including the database layer, the retrieval layer, the integration layer, and the adaptation layer. In the database layer, data acquired by multimodal sensors and sources from the dynamic urban environment are stored in the knowledge base. In addition, a tool set includes multi-domain functions which the urban system provides is available for retriever to get tool execution results. The retrieval layer dynamically queries a continually evolving knowledge base to extract task-specific information based on incoming urban inputs. The integration layer fuses the retrieved knowledge with model representations, enabling contextually informed reasoning. The adaptation layer incrementally updates the model parameters to incorporate new knowledge while preserving previously learned capabilities. Each layer is optimized with distinct objectives but coordinated under a unified multilevel optimization framework to maintain system-wide stability and adaptability.

The retrieval layer interfaces with a dynamic, continuously updated knowledge repository, which may include structured data e.g., urban maps, policy documents as well as unstructured data e.g., sensor feeds, social media reports. Retrieved knowledge is filtered and encoded into a format compatible with the language model's internal processing pipeline. The integration layer aligns this external information with internal contextual embeddings, allowing the model to ground its reasoning on both historical and newly acquired knowledge. The adaptation layer employs continual learning strategies, such as regularization and memory replay, to update model parameters while mitigating catastrophic forgetting. Together, these layers form a tightly coupled system capable of sustaining high-performance urban intelligence operations over long time horizons.

The knowledge retrieval pipeline in the UrbanMind framework is designed to dynamically extract relevant information from an evolving urban knowledge base. Upon receiving a query $x_t$ at time step t, the retrieval module first encodes the query into a latent representation using a lightweight encoder. This representation is then matched against indexed entries in the knowledge base $\mathcal{K}_t$ using similarity search techniques optimized for dynamic environments. To handle the heterogeneous nature of urban data, the knowledge base is organized into multiple modalities and domains, allowing the retrieval module to perform targeted, context-aware searches. Retrieved entries $\mathcal{R}_t(x_t)$ are filtered based on relevance scores and uncertainty estimates before being passed to the integration module for downstream processing.

Given the non-stationary nature of urban environments, the continual update of the knowledge base $\mathcal{K}_t$ is critical for maintaining retrieval accuracy and contextual relevance. New information streams, such as updated traffic reports, environmental sensor readings, or policy changes, are periodically ingest into $\mathcal{K}_t$ through an incremental indexing mechanism. Older entries are either updated or pruned based on criteria such as timestamp relevance, redundancy detection, and semantic consistency. To mitigate the risks of retrieval noise and knowledge drift, a validation layer monitors newly ingested entries, employing lightweight classifiers or rule-based filters to enforce minimal quality standards. This continual update mechanism ensures that the retrieval pipeline remains robust against concept shifts and information obsolescence.

The integration between continual retrieval and model adaptation is coordinated through a retrieval memory management module. This module maintains metadata regarding the retrieval history and past integration decisions, enabling the system to balance exploitation of stable historical knowledge and exploration of newly retrieved information. By dynamically adjusting retrieval strategies based on performance feedback, the framework ensures that knowledge integration remains both adaptive and stable. This continual coupling between retrieval updates and model adaptation forms the core mechanism that allows the UrbanMind system to achieve long-term resilience and effective decision-making in evolving urban environments.

\begin{figure}[h]
  \centering
  \includegraphics[width=0.8\linewidth,
                    trim=2cm 0cm 0cm 0cm,
                    clip                 
        ]{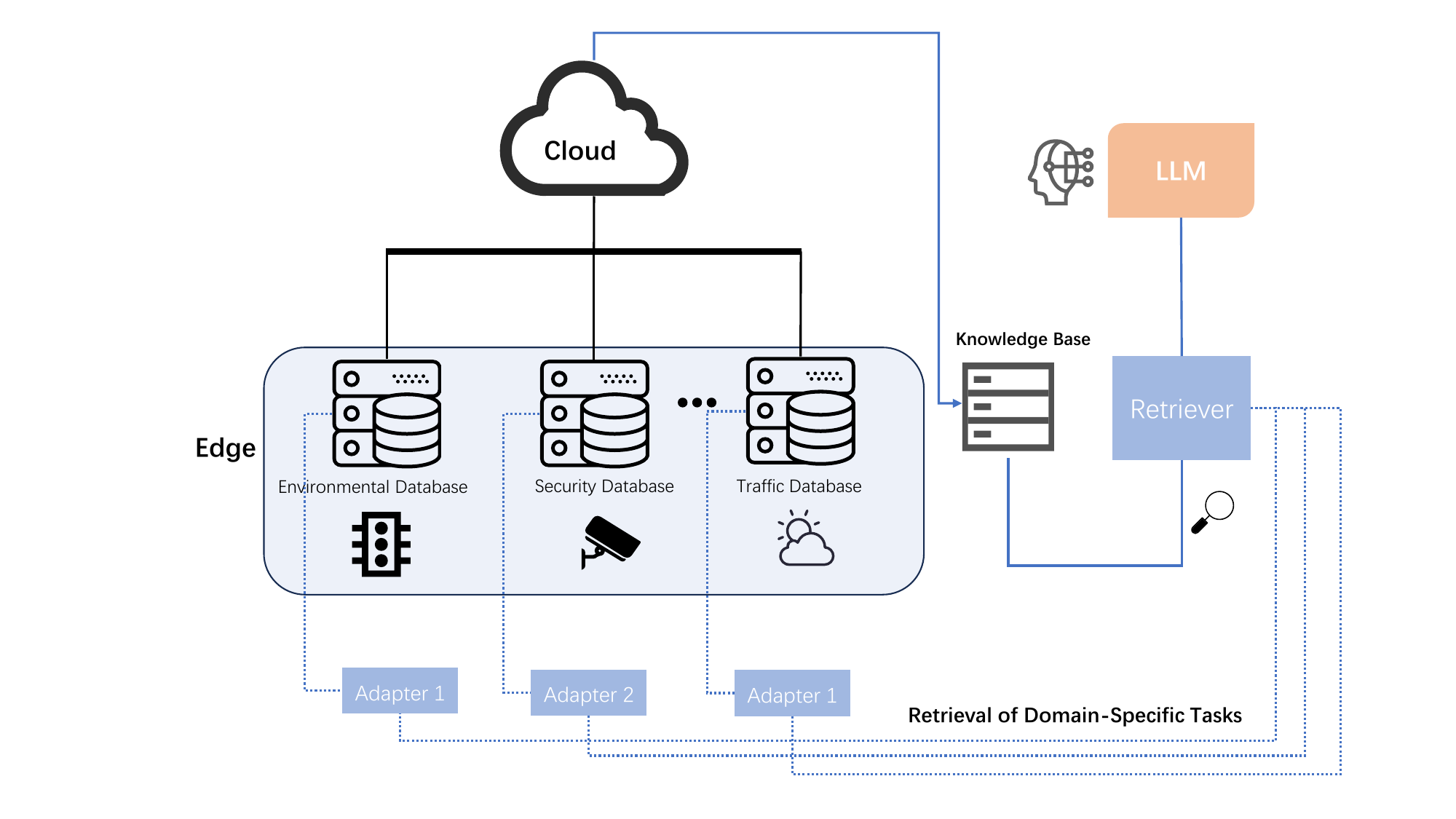}
  \caption{UrbanMind over Cloud-Edge Architecture}
  \Description{cloude-edge with UrbanMind}
  \label{fig:edgecloud with c-rag-llm}
\end{figure}

Notably, the proposed UrbanMind framework can be seamlessly implemented on a urban Cloud-Edge system \cite{di2025cloud} (Fig \ref{fig:edgecloud with c-rag-llm}), wherein the cloud layer is responsible for centralized orchestration and the management of global knowledge within LLM, while the edge layer focuses on localized data processing and personalized retrieval adaptation. Under this architecture, each edge node maintains an independent local knowledge base that captures region-specific and real-time information, such as traffic patterns or security surveillance data. Each edge database can be connected with a lightweight fine-tuning model, referred to as an adapter. The adapter is designed to enable efficient personalization and task adaptation through minimal parameter updates, without modifying the core parameters of the pre-trained language model. By allowing each edge node to train its adapter based on localized context and task-specific requirements, the framework supports the deployment of highly customized, context-aware intelligent services across heterogeneous urban environments.

\section{Multilevel Optimization Strategy for UrbanMind}

UrbanMind adopts a multilevel optimization strategy that aligns naturally with its modular design and integrates seamlessly with MoE architectures. This approach enables coordinated optimization across different layers of UrbanMind, allowing each component to specialize and adapt independently while maintaining overall system coherence. In addition, please note that this strategy is highly flexible, i.e., it supports both end-to-end training and targeted optimization of selected modules, making it suitable for a wide range of deployment scenarios and resource budgets. Its flexibility and generality ensure broad applicability across diverse urban tasks, providing robust and scalable performance in dynamic and data-driven environments.

\subsection{Multilevel Optimization}

Multilevel optimization is a hierarchical optimization framework in which the solution to an upper-level problem depends on the optimal solution of one or more lower-level problems. Multilevel optimization serves as a unifying framework that includes robust optimization \cite{yang2008distributed,yang2014distributed} and bilevel optimization \cite{jiao2022asynchronous} as special cases.   
\begin{equation}
\begin{aligned}
    & \min_{\bm{x}_1 \in \mathcal{X}_1, \bm{x}_2 \in \mathcal{X}_2, \ldots, \bm{x}_K \in \mathcal{X}_K} && \mathcal{F}_1(\bm{x}_1, \ldots, \bm{x}_K) \\
    & \text{s.t.} \quad && \bm{x}_2 \in \arg\min_{\bm{x}_2' \in \mathcal{X}_2(\bm{x}_1)} \mathcal{F}_2(\bm{x}_1, \bm{x}_2') \\
    & \quad && \bm{x}_3 \in \arg\min_{\bm{x}_3' \in \mathcal{X}_3(\bm{x}_1, \bm{x}_2)} \mathcal{F}_3(\bm{x}_1, \bm{x}_2, \bm{x}_3') \\
    & \quad && \vdots \\
    & \quad && \bm{x}_K \in \arg\min_{\bm{x}_K' \in \mathcal{X}_K(\bm{x}_1, \ldots, \bm{x}_{K-1})} \mathcal{F}_K(\bm{x}_1, \ldots, \bm{x}_{K-1}, \bm{x}_K'),
\end{aligned}
\end{equation}
where \(\bm{x}_1, \bm{x}_2, \ldots, \bm{x}_K\) represent decision variables at different hierarchical levels, each constrained by feasible sets \(\mathcal{X}_k\) that may depend on higher level variables. The objective functions \(\mathcal{F}_k\) correspond to each level’s objective and the lower-level problem's optimal as constraints. 

This structure naturally captures nested decision-making hierarchies, making it well suited for many real-world tasks with nested dependencies. In the field of machine learning, multilevel optimization has been widely adopted in various applications such as hyperparameter tuning, meta-learning \cite{kim2025stochastic}, and neural architecture search \cite{jiao2022timeautoad}, where it effectively captures the interplay between model training and evaluation \cite{liu2021investigating,jian2024tri}. Moreover, multilevel optimization can be implemented in a distributed manner \cite{jiao2022asynchronous,zhang2024multi}. 

Given its natural ability to model layered decision structures and integrate local and global objectives under uncertainty, multilevel optimization is particularly well suited for continual retrieval-augmented generation with large language models, where both the retrieval module and the generative model require joint, adaptive, and context-sensitive optimization over time. This is especially relevant in urban intelligence scenarios, where learning systems are inherently distributed across edge devices, sensors, and cloud infrastructure. In such settings, multilevel optimization provides a principled framework to coordinate decentralized learning, handle heterogeneous data sources, and adapt to dynamic environments in a scalable and robust manner.

\subsection{Dynamic Knowledge Retrieval Optimization}\label{subsec:dynamic_retrieval}

To maintain retrieval relevance in dynamic urban environments, the UrbanMind framework employs a task-aware retrieval strategy that adapts retrieval policies based on the current task context. At each time step $t$, given an input $x_t$ and its associated task descriptor $\tau_t$, the retrieval module dynamically selects a retrieval subspace within the knowledge base $\mathcal{K}_t$ that aligns with the semantic and operational requirements of $\tau_t$. Task descriptors are either explicitly provided e.g., transportation prediction or safety monitoring. By restricting retrieval to task-relevant domains, the system improves retrieval efficiency, reduces noise, and enhances the contextual grounding of the downstream tasks.

The retrieval scoring function is jointly optimized to account for both semantic similarity between $x_t$ and candidate knowledge entries, and task relevance based on $\tau_t$. Formally, the retrieval score $s(x_t, r_j; \tau_t)$ for a candidate entry $r_j$ is defined as a weighted combination of similarity metrics and task-specific relevance estimators. This composite scoring allows the retrieval module to favor entries that are not only lexically similar but also operationally significant for the target task. To adapt to evolving task definitions and domain shifts, the retrieval parameters are updated continually through feedback signals derived from downstream task performance.

Moreover, to ensure robustness under concept drift and multi-task scenarios, the task-aware retrieval strategy maintains a dynamic task profile memory. This memory captures historical retrieval patterns and associated task performance metrics, enabling the system to adjust retrieval subspace selection and scoring mechanisms over time. When encountering new tasks or unseen conditions, the system can leverage task memory to perform retrieval by analogy, drawing from similar prior experiences. By integrating task awareness into the retrieval process, the UrbanMind framework achieves greater flexibility, relevance, and resilience in knowledge acquisition for continually evolving urban intelligence applications.

In dynamic urban environments, the retrieval corpus $\mathcal{K}_t$ is subject to continuous evolution as new information becomes available and outdated information loses relevance. To maintain retrieval quality under data drift, the UrbanMind framework implements an incremental corpus update mechanism. New data streams such as sensor readings, event reports, or policy changes are continuously processed and indexed into $\mathcal{K}_t$ through a lightweight ingestion pipeline. Each incoming entry is associated with temporal metadata, task relevance scores, and uncertainty estimates to facilitate subsequent retrieval and maintenance decisions. To prevent uncontrolled corpus growth and semantic inconsistency, stale or low-relevance entries are periodically pruned based on temporal decay functions, redundancy detection, and performance-driven validation metrics.

Evaluating retrieval effectiveness in the UrbanMind framework requires metrics that capture both the relevance and robustness of retrieved knowledge under dynamic conditions. Standard retrieval metrics such as Top-k accuracy, Mean Reciprocal Rank (MRR), and Normalized Discounted Cumulative Gain (NDCG) are employed to measure how accurately the retrieved entries align with the ground-truth or task-specific information. In addition to static retrieval performance, continual settings necessitate tracking temporal retrieval stability, defined as the consistency of retrieval quality across evolving data distributions. Drift-aware metrics, such as relevance retention rate and retrieval degradation rate over time, are further utilized to quantify the system’s resilience to concept drift \cite{bayram2022concept}. Together, these metrics provide a comprehensive evaluation of retrieval performance in both stationary and non-stationary urban environments.

\subsection{Model Adaptation}\label{subsec:retrieval_conditioned_adaptation}

In the UrbanMind framework, the construction of retrieval-conditioned inputs is a critical step for effectively integrating external knowledge into the model's reasoning process. Upon retrieval, the selected entries $\mathcal{R}_t(x_t)$ are first processed through a knowledge encoding module that transforms heterogeneous data types, such as textual descriptions, sensor observations, or structured records, into unified latent representations. These encoded retrievals are then combined with the original input representation $h(x_t)$ through concatenation or fusion mechanisms specifically designed to preserve task-relevant information while minimizing redundancy. The retrieval-conditioned input, denoted as $\tilde{h}(x_t) = \mathcal{F}(h(x_t),\mathcal{R}_t(x_t))$, serves as the new context for downstream prediction or decision-making tasks.

To ensure that the retrieval-conditioned representations remain robust across dynamic environments, the fusion mechanism $\mathcal{F}(\cdot)$ is trained to selectively emphasize high-confidence, task-relevant knowledge while attenuating the influence of noisy or irrelevant retrievals. Attention-based weighting schemes, confidence scoring, and domain-specific gating functions are employed to dynamically modulate the contribution of each retrieved entry during integration. This retrieval-conditioned construction process enables the model to ground its reasoning not only on the immediate input but also on continually evolving external knowledge, providing a foundation for stable and adaptive learning in non-stationary urban environments.

To accommodate evolving urban knowledge and task distributions, the UrbanMind framework employs continual fine-tuning strategies conditioned on dynamically retrieved information. At each time step $t$, model updates are performed using retrieval-conditioned inputs $\tilde{h}(x_t)$ to align the model’s internal representations with the most recent knowledge context. Fine-tuning objectives incorporate regularization terms to preserve critical parameters associated with previous tasks, thereby mitigating catastrophic forgetting while allowing sufficient plasticity for adaptation. Dynamic sample selection mechanisms prioritize fine-tuning on high-confidence retrievals and task-critical examples, ensuring that model updates are both efficient and stability-preserving. Through this retrieval-aware continual fine-tuning process, the model incrementally refines its predictive capabilities in response to both input distribution shifts and knowledge base evolution.

\subsection{Multilevel Optimization for Urban LLMs Training}

Recent work leverages Mixture-of-Experts (MoE) architectures for LLMs, which activate only a small subset of expert networks per input, significantly reducing computation costs without major performance loss. Models like DeepSeek-R1 \cite{guo2025deepseek} and GLaM \cite{du2022glam} show MoE's effectiveness in complex tasks such as reasoning, code generation, and domain adaptation, while maintaining high efficiency.

The MoE framework typically consists of a \textit{gating network} $g(x; \theta_g)$ and a collection of \textit{experts} $\{e_i(x; \theta_{e_i})\}_{i=1}^N$. For a given input $x$, the gating function produces a sparse distribution over experts, activating only a few (e.g., top-1 or top-2) to process the input. The training of such a system involves two main objectives:

\begin{itemize}
    \item \textit{Expert-specific loss minimization.} Each expert $e_i$ is trained to minimize its task-specific loss when it is activated. Let $\mathcal{L}_{e_i}(x; \theta_{e_i})$ denote the loss incurred by expert $i$, then the total expert loss for an input $x$ is weighted by the gating score:
    \begin{equation}
        \mathcal{L}_{\text{expert}}(x) = \sum_{i=1}^N g(x; \theta_g)_i \cdot \mathcal{L}_{e_i}(x; \theta_{e_i}).
    \end{equation}

    \item \textit{Routing quality regularization.} The gating network itself is trained to produce useful, stable, and balanced routing decisions. This may involve auxiliary losses such as entropy regularization \cite{zareapoor2024efficient}, load balancing \cite{chen2022ta}, and sparsity constraints \cite{gale2023megablocks}. We denote this combined loss as:
    \begin{equation}
        \mathcal{L}_r(x; \theta_g).
    \end{equation}
\end{itemize}

From the multilevel optimization perspective, the training of MoE LLM can be naturally formulated as a bilevel optimization problem:
\begin{equation}
\begin{aligned}
    &\min_{\theta_g} \; \mathcal{L}_r(\theta_g) + \mathbb{E}_x \left[ \mathcal{L}_{\mathrm{upper}}(\theta_g, \theta_e^*(\theta_g)) \right]
\\
&\text{s.t.} \quad
\theta_e^*(\theta_g) = \arg\min_{\theta_e} \mathbb{E}_x \left[ \sum_{i=1}^N g(x; \theta_g)_i \cdot \mathcal{L}_{e_i}(x; \theta_{e_i}) \right],
\end{aligned}
\end{equation}

where, $\theta_g$ denotes the routing parameters optimized at the upper-level to minimize routing loss $\mathcal{L}_r$ and overall task loss $\mathcal{L}_{\mathrm{upper}}$, while the lower-level optimizes expert parameters $\theta_e = \{\theta_{e_i}\}$ to minimize their weighted task losses conditioned on the routing decisions.

This bilevel structure explicitly models the hierarchical interaction in MoE LLM training, such that routing decisions guide expert specialization, enabling efficient and scalable Urban Foundation Model training.

\subsection{Multi-timescale RAG Optimization}
In dynamic urban environments, RAG systems face substantial uncertainty due to frequent and diverse changes. These uncertainties occur across different timescales and arise from the non-stationarity of data distributions and the prevalence of OOD scenarios. Examples include seasonal traffic variations, infrastructure modifications, and unexpected events such as accidents or emergencies. To characterize such uncertainties, we introduce an uncertainty set \( \mathcal{U} \), representing potential distribution shifts. Then, we propose a multilevel formulation for multi-timescale end-to-end RAG optimization, aiming to ensure robust performance under worst-case scenarios.

To characterize the uncertainty set \(\mathcal{U}\), we employ divergence-based metrics to measure the shift between the empirical training distribution \(\mathcal{P}_{\text{train}}\) and possible test-time distributions. Common choices include Wasserstein distance \cite{ruschendorf1985wasserstein}, Jensen-Shannon divergence \cite{menendez1997jensen}, and \(L_2\) distance \cite{qian2019robust}, each offering different trade-offs in robustness and computational complexity. In this work, we adopt the Kullback-Leibler (KL) divergence as an example. Accordingly, the uncertainty set is defined as
$\mathcal{U} = \left\{ \mathcal{P}: D_{\text{KL}}(\mathcal{P} \| \mathcal{P}_{\text{train}}) \leq \rho \right\}$,
where \(\rho > 0\) controls the allowable shift from the training distribution.

The end-to-end optimization strategy of RAG in the UrbanMind framework is formulated as a bilevel optimization problem, as presented in \eqref{bilevel}. 

\begin{equation}
    \begin{aligned}
&\min_{\boldsymbol{\theta}} \mathcal{F}(\boldsymbol{\theta}, \boldsymbol{\phi}^*(\boldsymbol{\theta})) = \sum_{(q_j, a_j) \in \mathcal{D}_{\text{val}}} \ell_{\text{eval}}(a_j, \mathcal{G}(q_j, \mathcal{R}(q_j; \boldsymbol{\theta}); \boldsymbol{\phi}^*(\boldsymbol{\theta}))) \\
&\text{s.t.} \quad \boldsymbol{\phi}^*(\boldsymbol{\theta}) = \arg\min_{\boldsymbol{\phi}} \mathcal{L}_{\text{gen}}(\boldsymbol{\phi}; \boldsymbol{\theta}) = \sum_{(q_i, a_i) \in \mathcal{D}_{\text{train}}} \ell_{\text{train}}(a_i, \mathcal{G}(q_i, \mathcal{R}(q_i; \boldsymbol{\theta}); \boldsymbol{\phi})).
\end{aligned}
\label{bilevel}
\end{equation}

This formulation enables the joint optimization of the retriever \(\mathcal{R}\), parameterized by \(\boldsymbol{\theta}\), and the generator \(\mathcal{G}\), parameterized by \(\boldsymbol{\phi}\). Given a user query \(q_t\) at time step \(t\), the retriever selects a set of \(K\) documents, \(\mathcal{D}_t = \mathcal{R}(q_t; \boldsymbol{\theta}) = \{d_1, d_2, \ldots, d_K\}\), from the evolving knowledge base \(\mathcal{K}_t\). The generator then produces a response \(\hat{a}_t = \mathcal{G}(q_t, \mathcal{D}_t(q_t; \boldsymbol{\theta}); \boldsymbol{\phi})\), conditioned on the query and the retrieved documents.

From the perspective of Distributionally Robust Optimization (DRO) \cite{jiaoasynchronous}, the end-to-end formulation can be naturally extended to a multilevel optimization framework, where uncertainty in dynamic urban environments is explicitly modeled through the uncertainty set \(\mathcal{U}\). This approach provides a principled foundation for enhancing robustness to distributional shifts by optimizing model performance under worst-case scenarios within \(\mathcal{U}\). Suppose that the training dataset \(\mathcal{D}_{\text{train}}\) is composed of \(M\) domains (e.g., traffic data during peak hours, public safety data, urban planning data), denoted as \(\{\mathcal{D}_1, \mathcal{D}_2, \ldots, \mathcal{D}_M\}\). A weight vector \(\boldsymbol{w} = [w_1, w_2, \ldots, w_M]\) is introduced , where \(w_m\) represents the importance of domain \(\mathcal{D}_m\), which incorporates DRO to learn \(\boldsymbol{w}\), such that the model generalizes better across diverse and evolving distributions. The multilevel optimization problem is defined as follows:
\begin{equation}
    \begin{aligned}
&\min_{\boldsymbol{\theta}} \mathcal{F}(\boldsymbol{\theta}, \boldsymbol{\phi}^*(\boldsymbol{\theta}), \boldsymbol{w}^*(\boldsymbol{\theta})) = \sum_{(q_j, a_j) \in \mathcal{D}_{\text{val}}} \ell_{\text{val}}(a_j, \mathcal{G}(q_j, \mathcal{R}(q_j; \boldsymbol{\theta}); \boldsymbol{\phi}^*(\boldsymbol{\theta}))) \\
&\text{s.t.} \quad \boldsymbol{\phi}^*(\boldsymbol{\theta}) = \arg\min_{\boldsymbol{\phi}} \mathcal{L}_{\text{gen}}(\boldsymbol{\phi}; \boldsymbol{\theta}, \boldsymbol{w}^*(\boldsymbol{\theta})) = \sum_{m=1}^M w_m^* \sum_{(q_i, a_i) \in \mathcal{D}_m} \ell_{\text{train}}(a_i, \mathcal{G}(q_i, \mathcal{R}(q_i; \boldsymbol{\theta}); \boldsymbol{\phi})), \\
&\quad\quad\text{s.t.} \quad \boldsymbol{w}^*(\boldsymbol{\theta}) = \arg\min_{\boldsymbol{w}} \mathcal{L}_{\text{dro}}(\boldsymbol{w}; \boldsymbol{\theta}, \boldsymbol{\phi}) = \sum_{m=1}^M w_m \sum_{(q_i, a_i) \in \mathcal{D}_m} \ell_{\text{train}}(a_i, \mathcal{G}(q_i, \mathcal{R}(q_i; \boldsymbol{\theta}); \boldsymbol{\phi})), \\
&\quad\quad\quad\quad \text{s.t.} \quad \sum_{m=1}^M w_m = 1, \quad \forall m \in \{1, \ldots, M\}, \\
&\quad\quad\qquad\qquad\text{ }\text{KL}(\boldsymbol{w} \| \boldsymbol{p}_{\text{uniform}}) <\epsilon,
\end{aligned}
\end{equation}
where \(\mathcal{D}_{\text{train}}\) and \(\mathcal{D}_{\text{val}}\) denote the training and validation datasets, \(\ell_{\text{train}}\) and \(\ell_{\text{val}}\) are the training and evaluation losses, respectively, \(\lambda\) is a regularization coefficient, \(\boldsymbol{p}_{\text{uniform}}\) is the uniform distribution over domains, and \(\epsilon > 0\) is a small constant to prevent over-concentration of weights. The first-level problem optimizes the retriever \(\boldsymbol{\theta}\) to improve overall performance on the validation set. The second-level problem optimizes the generator \(\boldsymbol{\phi}\) using a weighted training loss, where the weights \(\boldsymbol{w}\) prioritize contributions from different domains. The third-level problem employs DRO to learn \(\boldsymbol{w}\), balancing domain contributions with a KL divergence regularization to ensure diversity and robustness against distributional shifts.

Please note that this multilevel formulation can be adjusted to various \emph{temporal scales of module updates} within the RAG framework. For instance, if only the retriever $\mathcal{R}$, parameterized by $\boldsymbol{\theta}$, is to be optimized while the generator $\mathcal{G}$ and domain weights $\boldsymbol{w}$ are assumed to be fixed (e.g., pre-trained or updated less frequently), then the overall optimization reduces to a \emph{single-level} problem with respect to $\boldsymbol{\theta}$. Similarly, when the generator parameters $\boldsymbol{\phi}$ are fixed, the optimization focuses on learning a \emph{robust retriever} under distributional shifts through the upper-level objective involving $\boldsymbol{\theta}$ and $\boldsymbol{w}$, effectively resulting in a bilevel problem.

In addition, under different update schedules or timescales for retriever and generator components, the original multilevel problem can be \emph{decomposed or relaxed} into a sequence of tractable sub-problems. Each sub-problem can be solved separately, enabling practical and modular training strategies aligned with the dynamic nature of continual learning in evolving urban environments.

A key challenge in this framework is the computational complexity of multilevel optimization, compounded by the non-differentiability of the retriever's output due to the discrete top-\(K\) document selection process. Additionally, to adapt to the evolving knowledge base \(\mathcal{K}_t\), the retriever incorporates task-aware scoring, as described in Section~\ref{subsec:dynamic_retrieval}, weighting document relevance based on task descriptors \(\tau_t\). This ensures contextual alignment with current urban intelligence tasks.

\section{Evaluations}

In this section, we present a systematic evaluation framework to rigorously assess the performance of proposed UrbanMind that enable a plethora of urban generative intelligence tasks \cite{zhao2024retrieval}. 

Level-1: Urban tasks focus on retrieving explicit factual information directly from available urban datasets without requiring complex reasoning or inference. These tasks involve identifying and extracting specific details such as traffic incident reports, public transportation schedules, air quality indices, or zoning regulations, which are explicitly recorded in structured or semi-structured data sources. For example, answering queries like "What is the current congestion level on Highway XXX?" or "Which zones are designated for residential use in the downtown area?" requires the system to locate and extract the relevant factual information without synthesizing or extrapolating beyond the provided data. 

Level-2: Urban tasks for implicit facts from available urban data, requiring basic logical reasoning or simple cross-referencing across multiple information sources. Unlike Level-1 tasks, where information is directly accessible, Level-2 tasks demand that the system perform elementary deductions or combine dispersed data segments to derive the correct answer. For example, answering a query such as "Which public transportation lines are most affected by the ongoing road construction near Central Avenue?" necessitates correlating information about construction zones with transit route maps and service updates. These tasks test the model's ability to integrate related factual data points and apply straightforward reasoning, thereby representing a critical step toward enabling more context-aware and intelligent urban decision-making.

Level 3: Urban tasks require not only the retrieval of factual information but also the comprehension and application of domain-specific complex rationales that govern decision-making within the urban context. For example, evaluating whether a proposed urban development complies with zoning regulations involves interpreting legal statutes, procedural workflows, and multi-step approval processes. Likewise, understanding emergency response prioritization across varying incident types may involve extracting implicit practices from historical dispatch and resolution logs.

\subsection{Experimental Setup}

To evaluate the performance of UrbanMind across the defined levels of urban tasks, we conducted experiments on a high-performance computing environment. The experiments were run on an Ubuntu 22.04 LTS server equipped with an NVIDIA RTX 4090 GPU (24GB VRAM), 128GB of DDR5 RAM, and an Intel Core i9-13900K CPU (24 cores, 32 threads). This setup ensured efficient processing of large-scale urban datasets and the computational demands of continual learning and retrieval-augmented generation.

We developed an interactive evaluation pipeline using the \textit{Streamlit} framework (version 1.32.0), which facilitated real-time user queries and visualization of model responses for urban intelligence tasks. The UrbanMind framework was implemented using \textit{PyTorch} (version 2.3.0) with CUDA 12.1 for GPU acceleration, and large model inference was optimized using \textit{vLLM} (version 0.6.3) to enhance throughput and reduce latency during evaluation. The retrieval component leveraged \textit{Milvus} (version 2.5.4) for fast similarity search over the dynamic knowledge base \(\mathcal{K}_t\) to handle structured urban data. Real-time data updates were simulated using synthetic streams generated from publicly available urban datasets \cite{geofabrik2025}.

\begin{figure}[htbp]
  \centering
  \begin{minipage}{0.49\linewidth}
    \centering
    \includegraphics[width=\linewidth,trim=190 10 90 0, clip]{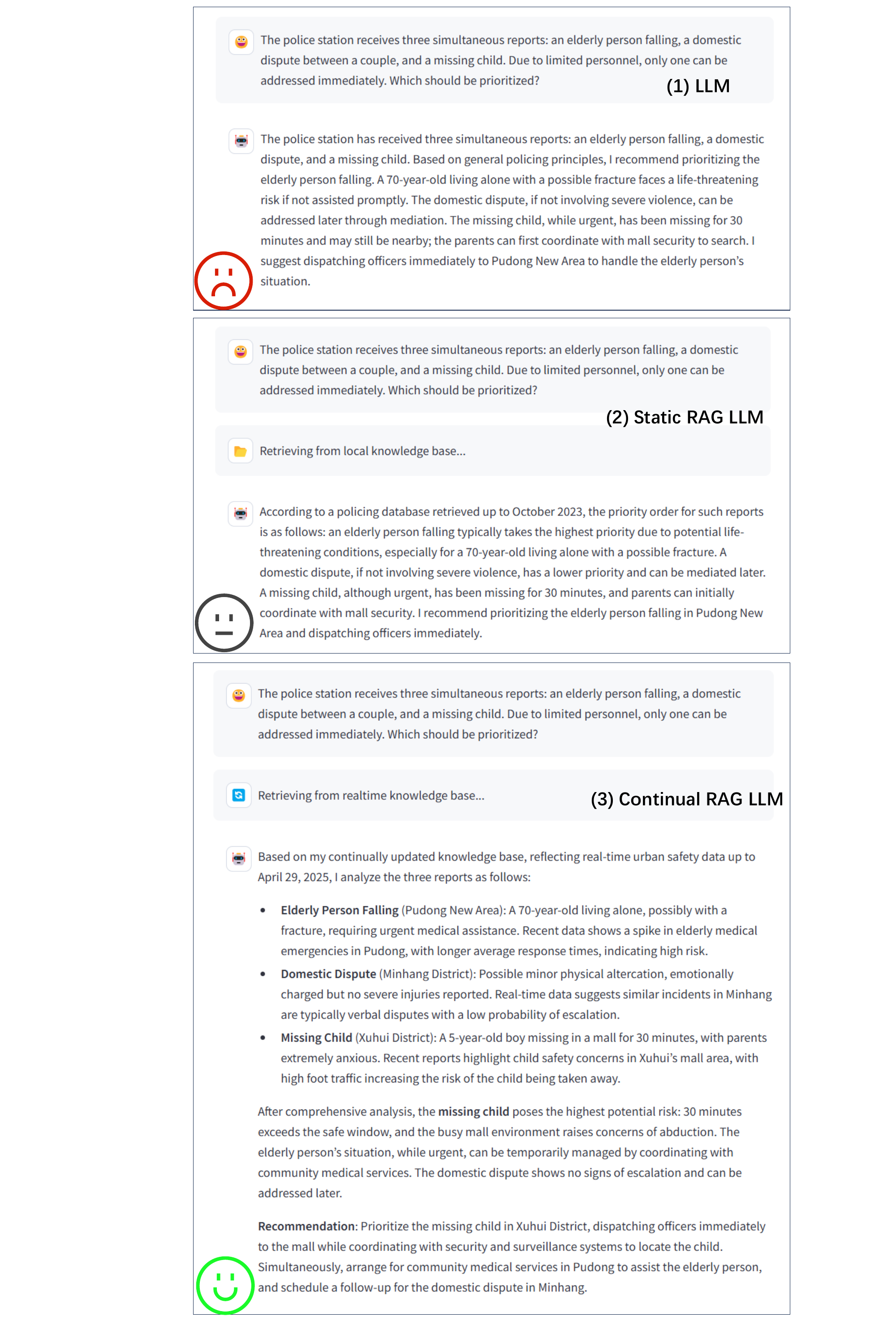} 
    \caption{LLM,Static RAG LLM and Continual RAG Comparison with Level 3 Task}
    \label{fig:level3}
  \end{minipage}
  \hfill
  \begin{minipage}{0.49\linewidth}
    \centering
    \includegraphics[width=\linewidth,,trim=50 50 30 40, clip]{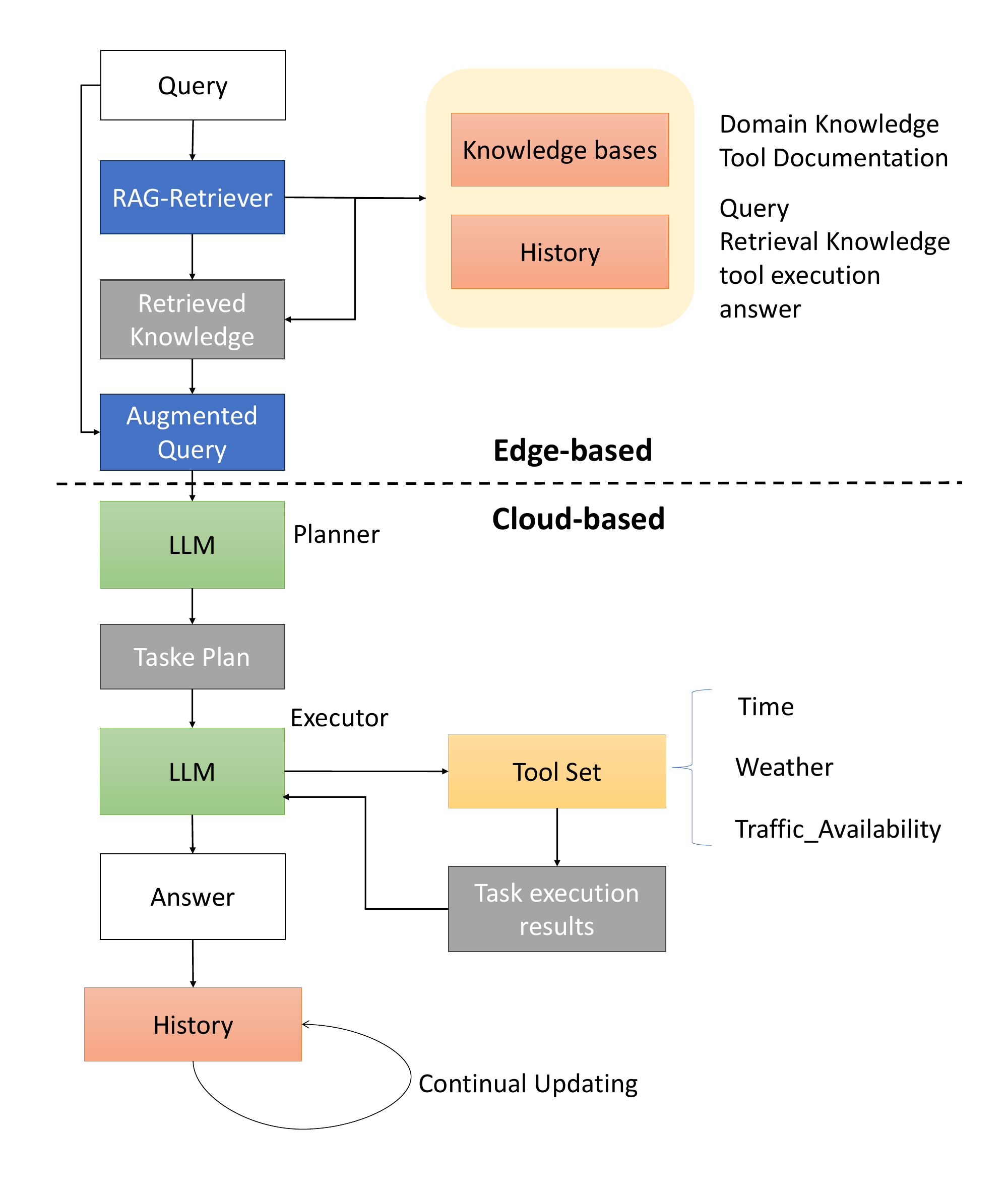}
    \caption{Tool-Enhanced UrbanMind Experimental Framework}
    \label{fig:tool_enhance_framework}
    
    \vspace{0.7em} 
    
    \includegraphics[width=\linewidth,trim=140 30 140 30, clip]{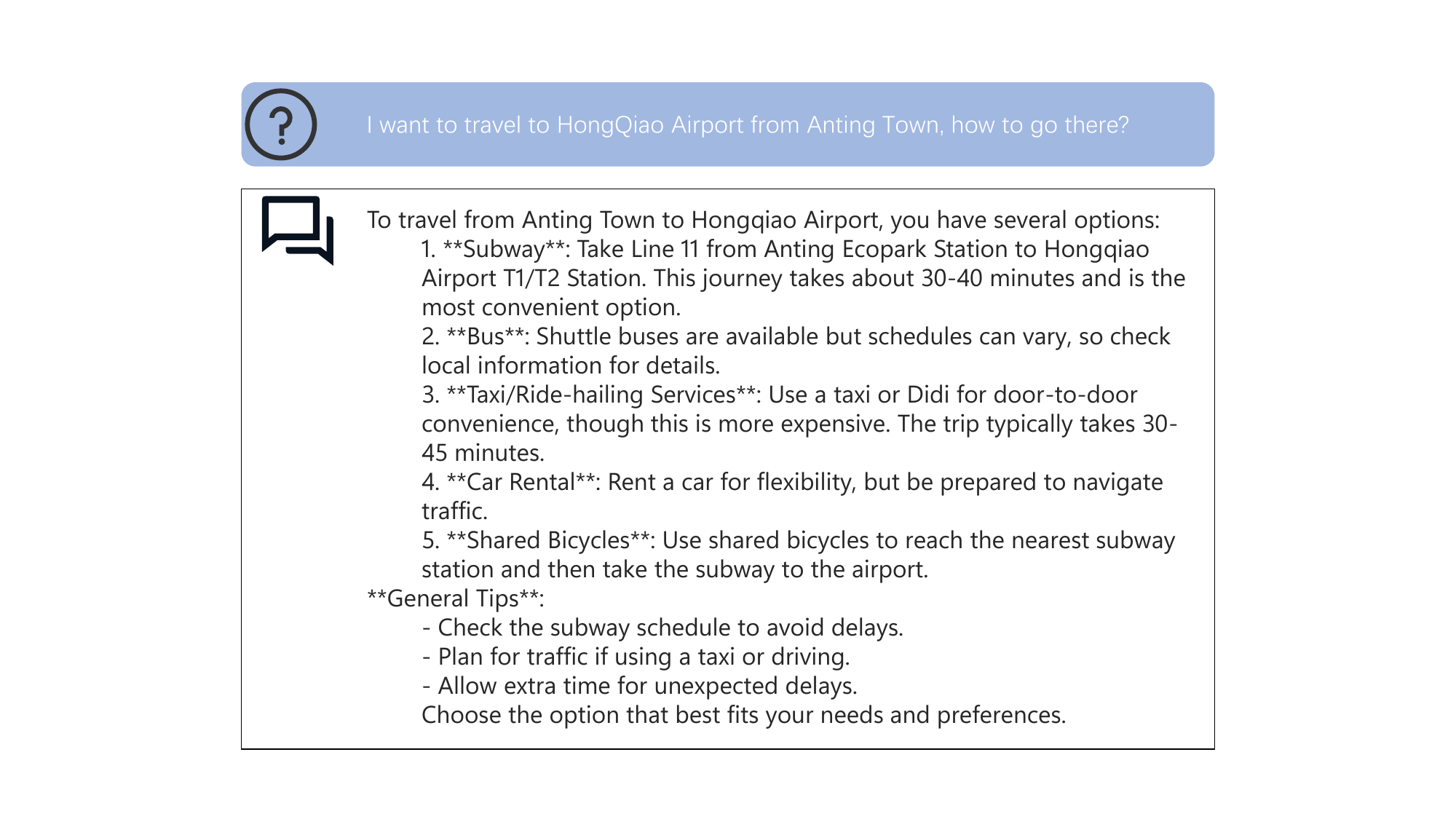} 
    \caption{Travel Plan with LLM-Only}
    \label{fig:llm_only}

  \end{minipage}
\end{figure}

To demonstrate the advantages of tool-enhanced RAG, including improved factual accuracy and support for multistep reasoning via domain-specific tool invocation, we develop a prototype of UrbanMind framework(Fig. \ref{fig:tool_enhance_framework}) for urban travel planning within a broader urban intelligence system. By integrating RAG with a modular tool-calling mechanism, the system dynamically selects and executes tools such as time and weather as well as traffic evaluators, based on user travel queries. The contextual information retrieved from the tool executions is incorporated into the LLM’s reasoning process, enabling adaptive planning of routes and transportation modes that account for real-time environmental constraints and urban dynamics. 

Specifically, in this tool-enhanced UrbanMind system, we constructed a toolset that includes a weather checking tool, a time tool, and a traffic availability access tool. A knowledge base with query-answer memory is maintained to store information about how to generate travel plans, the available tools, and interface documentation to support accurate tool invocation. The system is built using the LangChain \textit{PlanAndExecute} framework, leveraging the cloud-based \textit{Qwen2.5-32B-Instruct} model as the base LLM for enhanced generation capabilities, and a local PC-based RAG module for embedding and storing domain-specific knowledge.
\begin{figure}[htbp]
  \centering
  \begin{minipage}{0.49\linewidth}
    \centering
    \includegraphics[width=\linewidth,trim=190 370 90 190, clip]{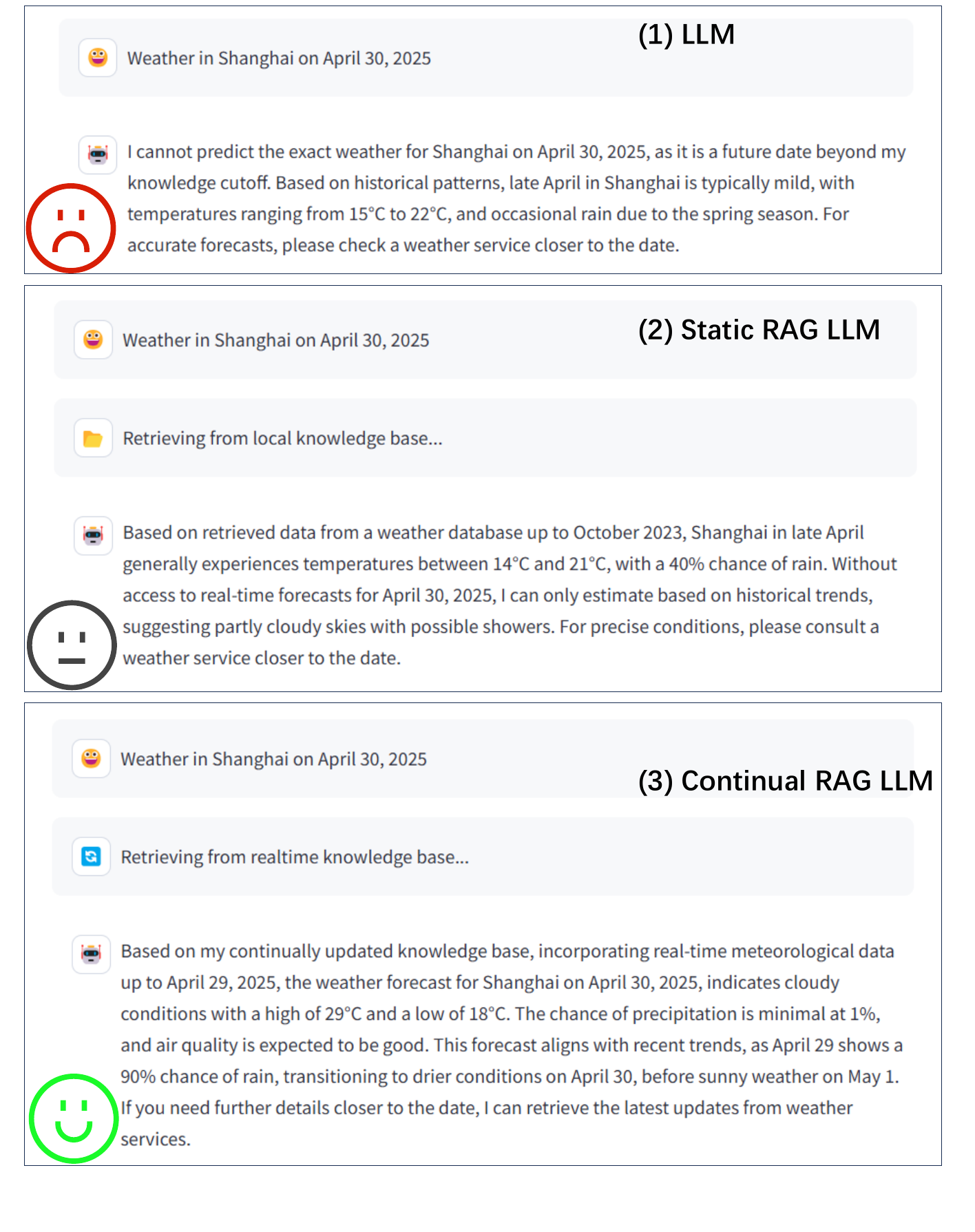} 
    \caption{LLM,Static RAG LLM and Continual RAG Comparison with Level 1 Task}
    \label{fig:level1}
  \end{minipage}
  \hfill
  \begin{minipage}{0.49\linewidth}
    \centering
    \includegraphics[width=\linewidth,,trim=190 370 90 190, clip]{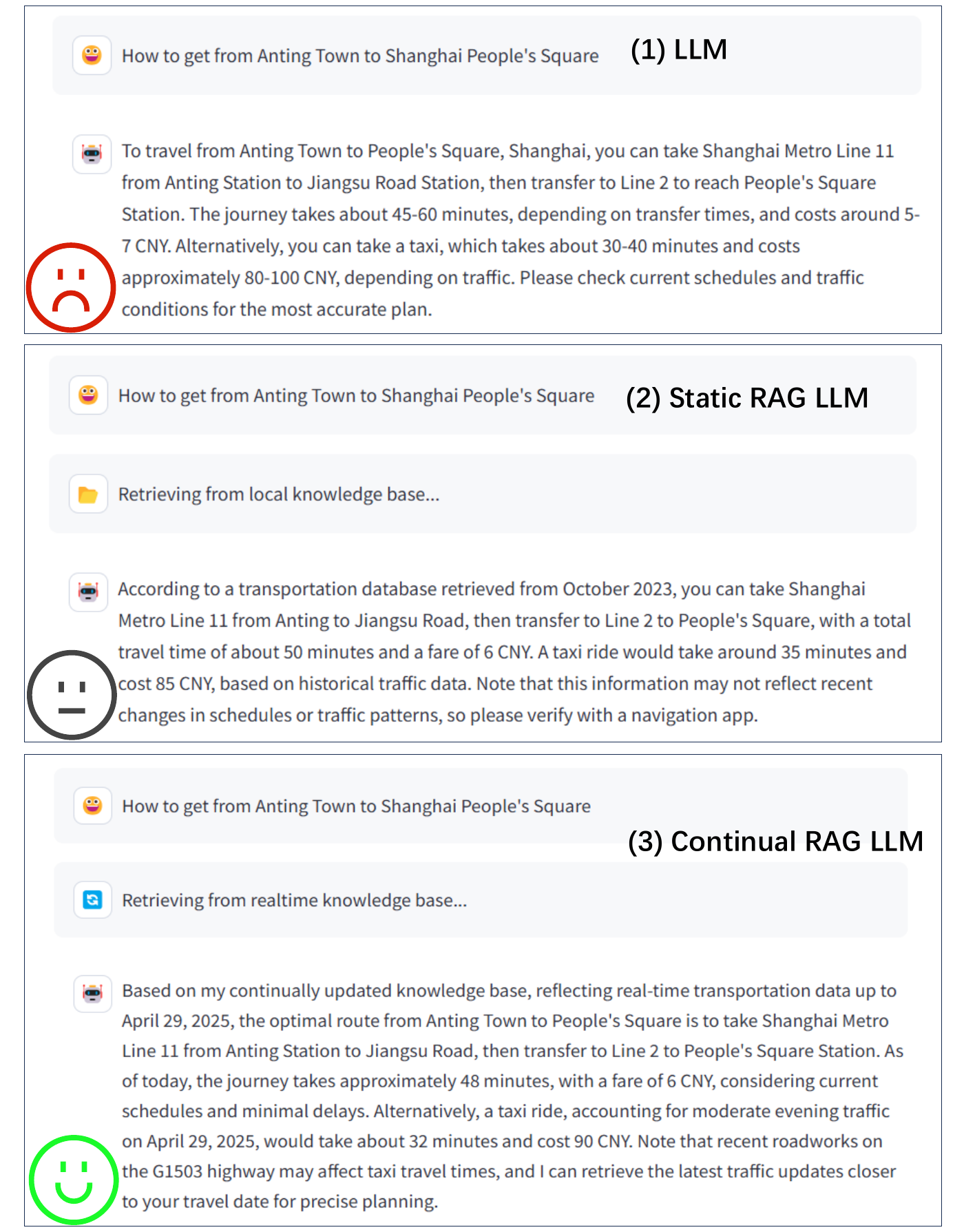}
    \caption{LLM,Static RAG LLM and Continual RAG Comparison with Level 2 Task}
    \label{fig:level2}
  \end{minipage}
\end{figure}

\begin{figure}[htbp]
  \centering
  \begin{minipage}{0.49\linewidth}
    \centering
    \includegraphics[width=\linewidth,trim=140 30 140 30, clip]{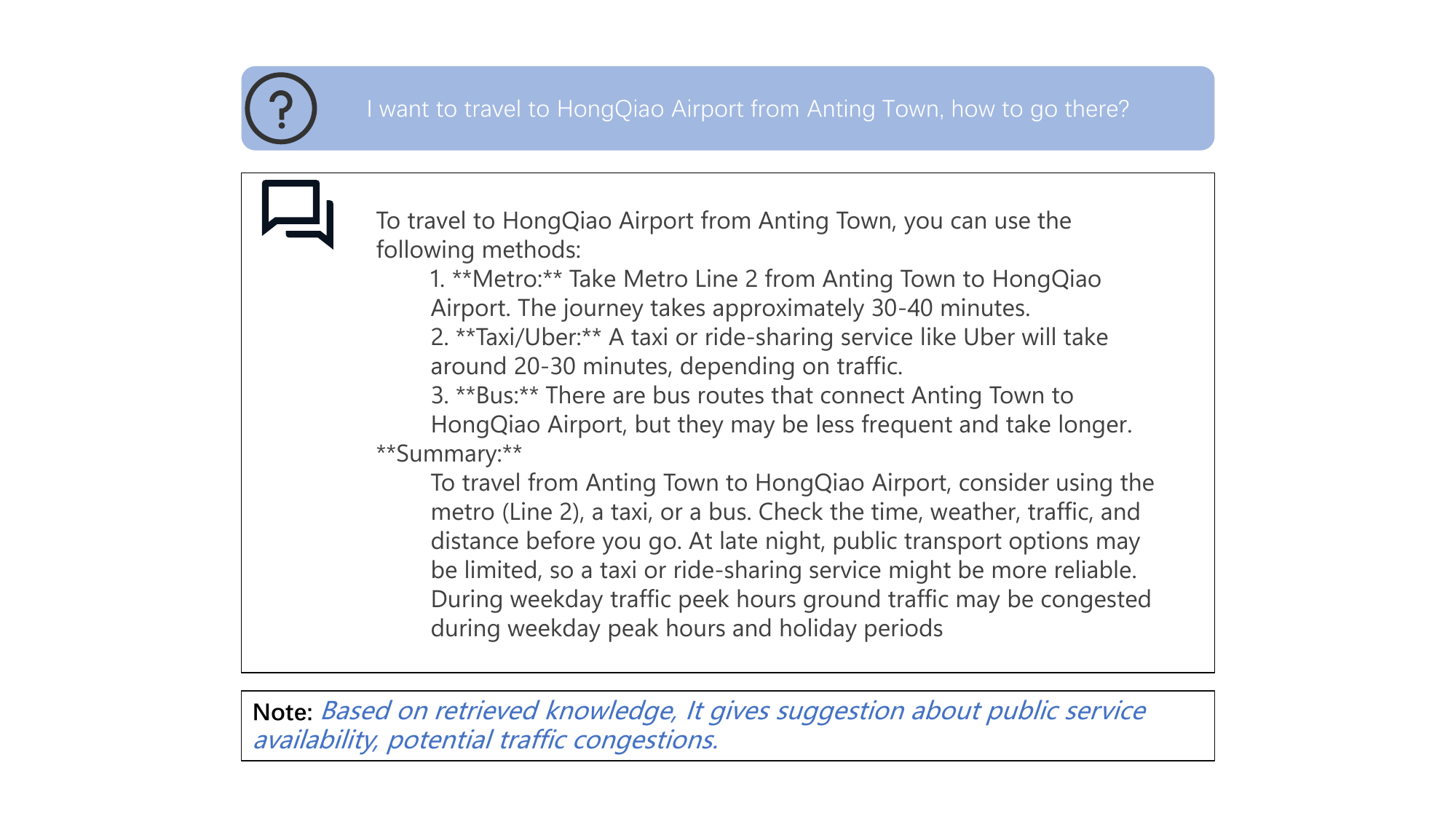} 
    \caption{Travel Plan with RAG-LLM}
    \label{fig:rag_llm}
  \end{minipage}
  \hfill
  \begin{minipage}{0.49\linewidth}
    \centering
    \includegraphics[width=\linewidth,,trim=140 30 140 30, clip]{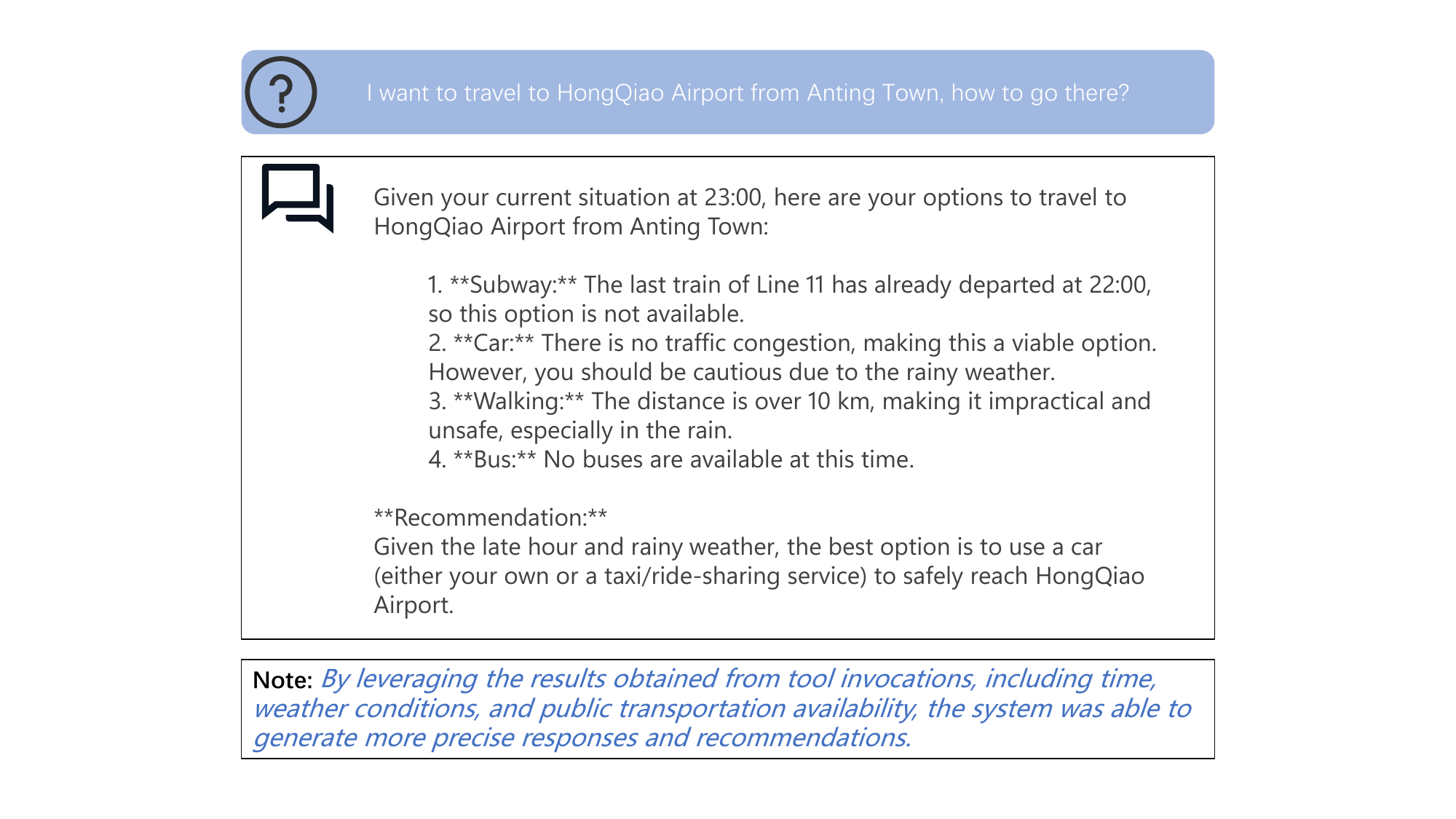}
    \caption{Travel Plan with Tool-enhanced UrbanMind}
    \label{fig:tool_enhanced_llm_result}
  \end{minipage}
\end{figure}
\subsection{Experimental Results}
Fig. \ref{fig:level1}, Fig. \ref{fig:level2}, and Fig. \ref{fig:level3} present the results of LLM-only generation, LLM with static RAG, and LLM with continual RAG, respectively, across the three task levels. The LLM-only approach lacks access to real-time information, while the static RAG model retrieves relevant prior experiences from the knowledge base to support the task. In contrast, the continual RAG model integrates up-to-date and time-sensitive data to provide the most accurate guidance. Across all task levels, the continual RAG-enhanced LLM consistently produces the most satisfactory responses, with its superiority in generation quality especially pronounced in lower-level tasks.

Fig. \ref{fig:llm_only}, Fig. \ref{fig:rag_llm}, and Fig. \ref{fig:tool_enhanced_llm_result} illustrate the travel planning outcomes generated by the LLM-only system, the RAG-LLM system, and the Tool-enhanced UrbanMind system, respectively. The LLM-only system is capable of suggesting general travel routes from location A to location B; however, it fails to incorporate contextual factors such as time and weather, thus limiting the relevance and accuracy of its recommendations. The RAG-LLM system, which augments the LLM with retrieved knowledge including public schedules and historical query-answer pairs, is able to offer more informative suggestions, for instance, checking weather conditions or public transport availability, as well as identifying potential traffic congestion. Nevertheless, it still lacks the ability to generate a coherent and concrete travel plan. In contrast, the Tool-enhanced UrbanMind system not only retrieves relevant knowledge through RAG but also enables the LLM to autonomously identify and invoke appropriate tools. Based on the real-time outputs of these tools, the system delivers a detailed and context-aware travel plan tailored to the user's query.

\section{Conclusion}

This paper introduces UrbanMind, a tool-enhanced RAG framework designed to advance urban general intelligence, which aims to enable urban intelligence system  to incrementally incorporate evolving urban data through corpus updating and supports privacy-preserving, low-latency inference via cloud-edge deployment. We further formulate the learning process of UrbanMind within a multilevel optimization framework that aligns naturally with the architecture of MoE LLMs. This formulation treats retrieval, generation, and model adaptation as interdependent subproblems, allowing for selective or multi-timescale end-to-end optimization according to resource constraints. Empirical evaluations on diverse real-world urban tasks demonstrate the effectiveness and versatility of the proposed framework. Collectively, UrbanMind marks a step toward realizing practical and adaptable UGI systems for future cities.


\bibliographystyle{ACM-Reference-Format}
\bibliography{reference}


\begin{thebibliography}{93}


\ifx \showCODEN    \undefined \def \showCODEN     #1{\unskip}     \fi
\ifx \showISBNx    \undefined \def \showISBNx     #1{\unskip}     \fi
\ifx \showISBNxiii \undefined \def \showISBNxiii  #1{\unskip}     \fi
\ifx \showISSN     \undefined \def \showISSN      #1{\unskip}     \fi
\ifx \showLCCN     \undefined \def \showLCCN      #1{\unskip}     \fi
\ifx \shownote     \undefined \def \shownote      #1{#1}          \fi
\ifx \showarticletitle \undefined \def \showarticletitle #1{#1}   \fi
\ifx \showURL      \undefined \def \showURL       {\relax}        \fi
\providecommand\bibfield[2]{#2}
\providecommand\bibinfo[2]{#2}
\providecommand\natexlab[1]{#1}
\providecommand\showeprint[2][]{arXiv:#2}

\bibitem[Abbes et~al\mbox{.}(2016)]%
        {abbes2016urban}
\bibfield{author}{\bibinfo{person}{Ali~Ben Abbes}, \bibinfo{person}{ImedRiadh Farah}, {and} \bibinfo{person}{Vincent Barra}.} \bibinfo{year}{2016}\natexlab{}.
\newblock \showarticletitle{Urban growth analysis using multi-temporal satellite images, non-stationary decomposition methods and stochastic modeling}.
\newblock \bibinfo{journal}{\emph{World Acad Sci Eng Technol Int J Comput Electr Autom Control Inf Eng}} \bibinfo{volume}{10}, \bibinfo{number}{10} (\bibinfo{year}{2016}), \bibinfo{pages}{1791--1797}.
\newblock


\bibitem[Ait~Ouallane et~al\mbox{.}(2022)]%
        {ait2022fusion}
\bibfield{author}{\bibinfo{person}{Asma Ait~Ouallane}, \bibinfo{person}{Assia Bakali}, \bibinfo{person}{Ayoub Bahnasse}, \bibinfo{person}{Said Broumi}, {and} \bibinfo{person}{Mohamed Talea}.} \bibinfo{year}{2022}\natexlab{}.
\newblock \showarticletitle{Fusion of engineering insights and emerging trends: Intelligent urban traffic management system}.
\newblock \bibinfo{journal}{\emph{Information Fusion}}  \bibinfo{volume}{88} (\bibinfo{year}{2022}), \bibinfo{pages}{218--248}.
\newblock


\bibitem[Ali and Wani(2024)]%
        {ali2024tri}
\bibfield{author}{\bibinfo{person}{Sarwat Ali} {and} \bibinfo{person}{M~Arif Wani}.} \bibinfo{year}{2024}\natexlab{}.
\newblock \showarticletitle{Tri-level Optimization for Gradient-based Neural Architecture Search}. In \bibinfo{booktitle}{\emph{2024 International Conference on Machine Learning and Applications (ICMLA)}}. IEEE, \bibinfo{pages}{1546--1552}.
\newblock


\bibitem[Balsebre et~al\mbox{.}(2023)]%
        {balsebre2023cityfm}
\bibfield{author}{\bibinfo{person}{Pasquale Balsebre}, \bibinfo{person}{Weiming Huang}, \bibinfo{person}{Gao Cong}, {and} \bibinfo{person}{Yi Li}.} \bibinfo{year}{2023}\natexlab{}.
\newblock \showarticletitle{Cityfm: City foundation models to solve urban challenges}.
\newblock \bibinfo{journal}{\emph{arXiv preprint arXiv:2310.00583}} (\bibinfo{year}{2023}).
\newblock


\bibitem[Bayram et~al\mbox{.}(2022)]%
        {bayram2022concept}
\bibfield{author}{\bibinfo{person}{Firas Bayram}, \bibinfo{person}{Bestoun~S Ahmed}, {and} \bibinfo{person}{Andreas Kassler}.} \bibinfo{year}{2022}\natexlab{}.
\newblock \showarticletitle{From concept drift to model degradation: An overview on performance-aware drift detectors}.
\newblock \bibinfo{journal}{\emph{Knowledge-Based Systems}}  \bibinfo{volume}{245} (\bibinfo{year}{2022}), \bibinfo{pages}{108632}.
\newblock


\bibitem[Bettencourt(2021)]%
        {bettencourt2021introduction}
\bibfield{author}{\bibinfo{person}{Lu{\'\i}s~MA Bettencourt}.} \bibinfo{year}{2021}\natexlab{}.
\newblock \showarticletitle{Introduction to urban science: evidence and theory of cities as complex systems}.
\newblock  (\bibinfo{year}{2021}).
\newblock


\bibitem[Bibri(2021)]%
        {bibri2021data}
\bibfield{author}{\bibinfo{person}{Simon~Elias Bibri}.} \bibinfo{year}{2021}\natexlab{}.
\newblock \showarticletitle{Data-driven smart sustainable cities of the future: Urban computing and intelligence for strategic, short-term, and joined-up planning}.
\newblock \bibinfo{journal}{\emph{Computational Urban Science}} \bibinfo{volume}{1}, \bibinfo{number}{1} (\bibinfo{year}{2021}), \bibinfo{pages}{8}.
\newblock


\bibitem[Biesialska et~al\mbox{.}(2020)]%
        {biesialska2020continual}
\bibfield{author}{\bibinfo{person}{Magdalena Biesialska}, \bibinfo{person}{Katarzyna Biesialska}, {and} \bibinfo{person}{Marta~R Costa-Jussa}.} \bibinfo{year}{2020}\natexlab{}.
\newblock \showarticletitle{Continual lifelong learning in natural language processing: A survey}.
\newblock \bibinfo{journal}{\emph{arXiv preprint arXiv:2012.09823}} (\bibinfo{year}{2020}).
\newblock


\bibitem[Chen et~al\mbox{.}(2022)]%
        {chen2022ta}
\bibfield{author}{\bibinfo{person}{Chang Chen}, \bibinfo{person}{Min Li}, \bibinfo{person}{Zhihua Wu}, \bibinfo{person}{Dianhai Yu}, {and} \bibinfo{person}{Chao Yang}.} \bibinfo{year}{2022}\natexlab{}.
\newblock \showarticletitle{Ta-moe: Topology-aware large scale mixture-of-expert training}.
\newblock \bibinfo{journal}{\emph{Advances in Neural Information Processing Systems}}  \bibinfo{volume}{35} (\bibinfo{year}{2022}), \bibinfo{pages}{22173--22186}.
\newblock


\bibitem[Chen et~al\mbox{.}(2021b)]%
        {chen2021trafficstream}
\bibfield{author}{\bibinfo{person}{Xu Chen}, \bibinfo{person}{Junshan Wang}, {and} \bibinfo{person}{Kunqing Xie}.} \bibinfo{year}{2021}\natexlab{b}.
\newblock \showarticletitle{TrafficStream: A streaming traffic flow forecasting framework based on graph neural networks and continual learning}.
\newblock \bibinfo{journal}{\emph{arXiv preprint arXiv:2106.06273}} (\bibinfo{year}{2021}).
\newblock


\bibitem[Chen et~al\mbox{.}(2024)]%
        {chen2024robust}
\bibfield{author}{\bibinfo{person}{Xingdi Chen}, \bibinfo{person}{Yu Xiong}, {and} \bibinfo{person}{Kai Yang}.} \bibinfo{year}{2024}\natexlab{}.
\newblock \showarticletitle{Robust beamforming for downlink multi-cell systems: A bilevel optimization perspective}. In \bibinfo{booktitle}{\emph{Proceedings of the AAAI Conference on Artificial Intelligence}}, Vol.~\bibinfo{volume}{38}. \bibinfo{pages}{7969--7977}.
\newblock


\bibitem[Chen et~al\mbox{.}(2021a)]%
        {chen2021intelligent}
\bibfield{author}{\bibinfo{person}{Zhuo Chen}, \bibinfo{person}{Ruoxi Chen}, {and} \bibinfo{person}{Songtao Chen}.} \bibinfo{year}{2021}\natexlab{a}.
\newblock \showarticletitle{Intelligent management information system of urban planning based on GIS}.
\newblock \bibinfo{journal}{\emph{Journal of Intelligent \& Fuzzy Systems}} \bibinfo{volume}{40}, \bibinfo{number}{4} (\bibinfo{year}{2021}), \bibinfo{pages}{6007--6016}.
\newblock


\bibitem[Devlin et~al\mbox{.}(2019)]%
        {devlin2019bert}
\bibfield{author}{\bibinfo{person}{Jacob Devlin}, \bibinfo{person}{Ming-Wei Chang}, \bibinfo{person}{Kenton Lee}, {and} \bibinfo{person}{Kristina Toutanova}.} \bibinfo{year}{2019}\natexlab{}.
\newblock \showarticletitle{Bert: Pre-training of deep bidirectional transformers for language understanding}. In \bibinfo{booktitle}{\emph{Proceedings of the 2019 conference of the North American chapter of the association for computational linguistics: human language technologies, volume 1 (long and short papers)}}. \bibinfo{pages}{4171--4186}.
\newblock


\bibitem[di~Martino et~al\mbox{.}(2025)]%
        {di2025cloud}
\bibfield{author}{\bibinfo{person}{Beniamino di Martino}, \bibinfo{person}{Domenico Di~Sivo}, {and} \bibinfo{person}{Alba Amato}.} \bibinfo{year}{2025}\natexlab{}.
\newblock \showarticletitle{Cloud, Edge, and Mobile Computing: Synergies for the Future of Smart Cities}. In \bibinfo{booktitle}{\emph{International Conference on Advanced Information Networking and Applications}}. Springer, \bibinfo{pages}{158--166}.
\newblock


\bibitem[Ding et~al\mbox{.}(2024)]%
        {ding2024realgen}
\bibfield{author}{\bibinfo{person}{Wenhao Ding}, \bibinfo{person}{Yulong Cao}, \bibinfo{person}{Ding Zhao}, \bibinfo{person}{Chaowei Xiao}, {and} \bibinfo{person}{Marco Pavone}.} \bibinfo{year}{2024}\natexlab{}.
\newblock \showarticletitle{Realgen: Retrieval augmented generation for controllable traffic scenarios}. In \bibinfo{booktitle}{\emph{European Conference on Computer Vision}}. Springer, \bibinfo{pages}{93--110}.
\newblock


\bibitem[Dou et~al\mbox{.}(2023)]%
        {dou2023towards}
\bibfield{author}{\bibinfo{person}{Fei Dou}, \bibinfo{person}{Jin Ye}, \bibinfo{person}{Geng Yuan}, \bibinfo{person}{Qin Lu}, \bibinfo{person}{Wei Niu}, \bibinfo{person}{Haijian Sun}, \bibinfo{person}{Le Guan}, \bibinfo{person}{Guoyu Lu}, \bibinfo{person}{Gengchen Mai}, \bibinfo{person}{Ninghao Liu}, {et~al\mbox{.}}} \bibinfo{year}{2023}\natexlab{}.
\newblock \showarticletitle{Towards artificial general intelligence (agi) in the internet of things (iot): Opportunities and challenges}.
\newblock \bibinfo{journal}{\emph{arXiv preprint arXiv:2309.07438}} (\bibinfo{year}{2023}).
\newblock


\bibitem[Dou et~al\mbox{.}(2024)]%
        {dou2024anomaly}
\bibfield{author}{\bibinfo{person}{Shaoyu Dou}, \bibinfo{person}{Kai Yang}, \bibinfo{person}{Yang Jiao}, \bibinfo{person}{Chengbo Qiu}, {and} \bibinfo{person}{Kui Ren}.} \bibinfo{year}{2024}\natexlab{}.
\newblock \showarticletitle{Anomaly Detection in Event-triggered Traffic Time Series via Similarity Learning}.
\newblock \bibinfo{journal}{\emph{IEEE Transactions on Dependable and Secure Computing}} (\bibinfo{year}{2024}).
\newblock


\bibitem[Du et~al\mbox{.}(2022)]%
        {du2022glam}
\bibfield{author}{\bibinfo{person}{Nan Du}, \bibinfo{person}{Yanping Huang}, \bibinfo{person}{Andrew~M Dai}, \bibinfo{person}{Simon Tong}, \bibinfo{person}{Dmitry Lepikhin}, \bibinfo{person}{Yuanzhong Xu}, \bibinfo{person}{Maxim Krikun}, \bibinfo{person}{Yanqi Zhou}, \bibinfo{person}{Adams~Wei Yu}, \bibinfo{person}{Orhan Firat}, {et~al\mbox{.}}} \bibinfo{year}{2022}\natexlab{}.
\newblock \showarticletitle{Glam: Efficient scaling of language models with mixture-of-experts}. In \bibinfo{booktitle}{\emph{International conference on machine learning}}. PMLR, \bibinfo{pages}{5547--5569}.
\newblock


\bibitem[Duran-Mateluna et~al\mbox{.}(2025)]%
        {duran2025adaptive}
\bibfield{author}{\bibinfo{person}{Cristian Duran-Mateluna} {et~al\mbox{.}}} \bibinfo{year}{2025}\natexlab{}.
\newblock \showarticletitle{Adaptive Robust Optimization Models for DER Planning in Distribution Networks under Long-and Short-Term Uncertainties}.
\newblock \bibinfo{journal}{\emph{arXiv e-prints}} (\bibinfo{year}{2025}), \bibinfo{pages}{arXiv--2503}.
\newblock


\bibitem[Elizalde-Ram{\'\i}rez et~al\mbox{.}(2019)]%
        {elizalde2019travel}
\bibfield{author}{\bibinfo{person}{Fernando Elizalde-Ram{\'\i}rez}, \bibinfo{person}{Romeo~Sanchez Nigenda}, \bibinfo{person}{Iris~A Mart{\'\i}nez-Salazar}, {and} \bibinfo{person}{Yasm{\'\i}n~{\'A} R{\'\i}os-Sol{\'\i}s}.} \bibinfo{year}{2019}\natexlab{}.
\newblock \showarticletitle{Travel plans in public transit networks using artificial intelligence planning models}.
\newblock \bibinfo{journal}{\emph{Applied Artificial Intelligence}} \bibinfo{volume}{33}, \bibinfo{number}{5} (\bibinfo{year}{2019}), \bibinfo{pages}{440--461}.
\newblock


\bibitem[Fadhel et~al\mbox{.}(2024)]%
        {fadhel2024comprehensive}
\bibfield{author}{\bibinfo{person}{Mohammed~A Fadhel}, \bibinfo{person}{Ali~M Duhaim}, \bibinfo{person}{Ahmed Saihood}, \bibinfo{person}{Ahmed Sewify}, \bibinfo{person}{Mokhaled~NA Al-Hamadani}, \bibinfo{person}{AS Albahri}, \bibinfo{person}{Laith Alzubaidi}, \bibinfo{person}{Ashish Gupta}, \bibinfo{person}{Sayedali Mirjalili}, {and} \bibinfo{person}{Yuantong Gu}.} \bibinfo{year}{2024}\natexlab{}.
\newblock \showarticletitle{Comprehensive systematic review of information fusion methods in smart cities and urban environments}.
\newblock \bibinfo{journal}{\emph{Information Fusion}} (\bibinfo{year}{2024}), \bibinfo{pages}{102317}.
\newblock


\bibitem[Fan et~al\mbox{.}(2025)]%
        {fan2025research}
\bibfield{author}{\bibinfo{person}{Yuxin Fan}, \bibinfo{person}{Yuxiang Wang}, \bibinfo{person}{Lipeng Liu}, \bibinfo{person}{Xirui Tang}, \bibinfo{person}{Na Sun}, {and} \bibinfo{person}{Zidong Yu}.} \bibinfo{year}{2025}\natexlab{}.
\newblock \showarticletitle{Research on the Online Update Method for Retrieval-Augmented Generation (RAG) Model with Incremental Learning}.
\newblock \bibinfo{journal}{\emph{arXiv preprint arXiv:2501.07063}} (\bibinfo{year}{2025}).
\newblock


\bibitem[Fei et~al\mbox{.}(2022)]%
        {fei2022towards}
\bibfield{author}{\bibinfo{person}{Nanyi Fei}, \bibinfo{person}{Zhiwu Lu}, \bibinfo{person}{Yizhao Gao}, \bibinfo{person}{Guoxing Yang}, \bibinfo{person}{Yuqi Huo}, \bibinfo{person}{Jingyuan Wen}, \bibinfo{person}{Haoyu Lu}, \bibinfo{person}{Ruihua Song}, \bibinfo{person}{Xin Gao}, \bibinfo{person}{Tao Xiang}, {et~al\mbox{.}}} \bibinfo{year}{2022}\natexlab{}.
\newblock \showarticletitle{Towards artificial general intelligence via a multimodal foundation model}.
\newblock \bibinfo{journal}{\emph{Nature Communications}} \bibinfo{volume}{13}, \bibinfo{number}{1} (\bibinfo{year}{2022}), \bibinfo{pages}{3094}.
\newblock


\bibitem[Floridi and Chiriatti(2020)]%
        {floridi2020gpt}
\bibfield{author}{\bibinfo{person}{Luciano Floridi} {and} \bibinfo{person}{Massimo Chiriatti}.} \bibinfo{year}{2020}\natexlab{}.
\newblock \showarticletitle{GPT-3: Its nature, scope, limits, and consequences}.
\newblock \bibinfo{journal}{\emph{Minds and Machines}}  \bibinfo{volume}{30} (\bibinfo{year}{2020}), \bibinfo{pages}{681--694}.
\newblock


\bibitem[Fortin et~al\mbox{.}(2021)]%
        {fortin2021use}
\bibfield{author}{\bibinfo{person}{Francis Fortin}, \bibinfo{person}{Julie Delle~Donne}, {and} \bibinfo{person}{Justine Knop}.} \bibinfo{year}{2021}\natexlab{}.
\newblock \showarticletitle{The use of social media in intelligence and its impact on police work}.
\newblock \bibinfo{journal}{\emph{Policing in an Age of Reform: An Agenda for Research and Practice}} (\bibinfo{year}{2021}), \bibinfo{pages}{213--231}.
\newblock


\bibitem[Freire et~al\mbox{.}(2021)]%
        {freire2021artificial}
\bibfield{author}{\bibinfo{person}{C{\'a}tia~AR Freire}, \bibinfo{person}{Fernando~AF Ferreira}, \bibinfo{person}{Elias~G Carayannis}, {and} \bibinfo{person}{Jo{\~a}o~JM Ferreira}.} \bibinfo{year}{2021}\natexlab{}.
\newblock \showarticletitle{Artificial intelligence and smart cities: A DEMATEL approach to adaptation challenges and initiatives}.
\newblock \bibinfo{journal}{\emph{IEEE Transactions on Engineering Management}} \bibinfo{volume}{70}, \bibinfo{number}{5} (\bibinfo{year}{2021}), \bibinfo{pages}{1881--1899}.
\newblock


\bibitem[Gale et~al\mbox{.}(2023)]%
        {gale2023megablocks}
\bibfield{author}{\bibinfo{person}{Trevor Gale}, \bibinfo{person}{Deepak Narayanan}, \bibinfo{person}{Cliff Young}, {and} \bibinfo{person}{Matei Zaharia}.} \bibinfo{year}{2023}\natexlab{}.
\newblock \showarticletitle{Megablocks: Efficient sparse training with mixture-of-experts}.
\newblock \bibinfo{journal}{\emph{Proceedings of Machine Learning and Systems}}  \bibinfo{volume}{5} (\bibinfo{year}{2023}), \bibinfo{pages}{288--304}.
\newblock


\bibitem[Gammelli et~al\mbox{.}(2023)]%
        {gammelli2023graph}
\bibfield{author}{\bibinfo{person}{Daniele Gammelli}, \bibinfo{person}{James Harrison}, \bibinfo{person}{Kaidi Yang}, \bibinfo{person}{Marco Pavone}, \bibinfo{person}{Filipe Rodrigues}, {and} \bibinfo{person}{Francisco~C Pereira}.} \bibinfo{year}{2023}\natexlab{}.
\newblock \showarticletitle{Graph reinforcement learning for network control via bi-level optimization}.
\newblock \bibinfo{journal}{\emph{arXiv preprint arXiv:2305.09129}} (\bibinfo{year}{2023}).
\newblock


\bibitem[Gao et~al\mbox{.}(2022)]%
        {gao2022efficient}
\bibfield{author}{\bibinfo{person}{Qiang Gao}, \bibinfo{person}{Zhipeng Luo}, \bibinfo{person}{Diego Klabjan}, {and} \bibinfo{person}{Fengli Zhang}.} \bibinfo{year}{2022}\natexlab{}.
\newblock \showarticletitle{Efficient architecture search for continual learning}.
\newblock \bibinfo{journal}{\emph{IEEE Transactions on Neural Networks and Learning Systems}} \bibinfo{volume}{34}, \bibinfo{number}{11} (\bibinfo{year}{2022}), \bibinfo{pages}{8555--8565}.
\newblock


\bibitem[{GeoFabrik GmbH}(2025)]%
        {geofabrik2025}
\bibfield{author}{\bibinfo{person}{{GeoFabrik GmbH}}.} \bibinfo{year}{2025}\natexlab{}.
\newblock \bibinfo{title}{{OpenStreetMap Data Extracts for China}}.
\newblock \bibinfo{howpublished}{\url{https://download.geofabrik.de/asia/china.html}}.
\newblock
\newblock
\shownote{Accessed: 2025-04-29}.


\bibitem[Goertzel(2014)]%
        {goertzel2014artificial}
\bibfield{author}{\bibinfo{person}{Ben Goertzel}.} \bibinfo{year}{2014}\natexlab{}.
\newblock \showarticletitle{Artificial general intelligence: concept, state of the art, and future prospects}.
\newblock \bibinfo{journal}{\emph{Journal of Artificial General Intelligence}} \bibinfo{volume}{5}, \bibinfo{number}{1} (\bibinfo{year}{2014}), \bibinfo{pages}{1}.
\newblock


\bibitem[Golpayegani et~al\mbox{.}(2021)]%
        {golpayegani2021urban}
\bibfield{author}{\bibinfo{person}{Fatemeh Golpayegani}, \bibinfo{person}{Saeedeh Ghanadbashi}, {and} \bibinfo{person}{Maha Riad}.} \bibinfo{year}{2021}\natexlab{}.
\newblock \showarticletitle{Urban emergency management using intelligent traffic systems: challenges and future directions}. In \bibinfo{booktitle}{\emph{2021 IEEE International Smart Cities Conference (ISC2)}}. IEEE, \bibinfo{pages}{1--4}.
\newblock


\bibitem[Guo et~al\mbox{.}(2025)]%
        {guo2025deepseek}
\bibfield{author}{\bibinfo{person}{Daya Guo}, \bibinfo{person}{Dejian Yang}, \bibinfo{person}{Haowei Zhang}, \bibinfo{person}{Junxiao Song}, \bibinfo{person}{Ruoyu Zhang}, \bibinfo{person}{Runxin Xu}, \bibinfo{person}{Qihao Zhu}, \bibinfo{person}{Shirong Ma}, \bibinfo{person}{Peiyi Wang}, \bibinfo{person}{Xiao Bi}, {et~al\mbox{.}}} \bibinfo{year}{2025}\natexlab{}.
\newblock \showarticletitle{Deepseek-r1: Incentivizing reasoning capability in llms via reinforcement learning}.
\newblock \bibinfo{journal}{\emph{arXiv preprint arXiv:2501.12948}} (\bibinfo{year}{2025}).
\newblock


\bibitem[Han et~al\mbox{.}(2024)]%
        {han2024llm}
\bibfield{author}{\bibinfo{person}{Shanshan Han}, \bibinfo{person}{Qifan Zhang}, \bibinfo{person}{Yuhang Yao}, \bibinfo{person}{Weizhao Jin}, \bibinfo{person}{Zhaozhuo Xu}, {and} \bibinfo{person}{Chaoyang He}.} \bibinfo{year}{2024}\natexlab{}.
\newblock \showarticletitle{LLM multi-agent systems: Challenges and open problems}.
\newblock \bibinfo{journal}{\emph{arXiv preprint arXiv:2402.03578}} (\bibinfo{year}{2024}).
\newblock


\bibitem[Hei et~al\mbox{.}(2024)]%
        {hei2024dr}
\bibfield{author}{\bibinfo{person}{Zijian Hei}, \bibinfo{person}{Weiling Liu}, \bibinfo{person}{Wenjie Ou}, \bibinfo{person}{Juyi Qiao}, \bibinfo{person}{Junming Jiao}, \bibinfo{person}{Guowen Song}, \bibinfo{person}{Ting Tian}, {and} \bibinfo{person}{Yi Lin}.} \bibinfo{year}{2024}\natexlab{}.
\newblock \showarticletitle{Dr-rag: Applying dynamic document relevance to retrieval-augmented generation for question-answering}.
\newblock \bibinfo{journal}{\emph{arXiv preprint arXiv:2406.07348}} (\bibinfo{year}{2024}).
\newblock


\bibitem[Huang et~al\mbox{.}(2024)]%
        {huang2024tool}
\bibfield{author}{\bibinfo{person}{Zhongzhen Huang}, \bibinfo{person}{Kui Xue}, \bibinfo{person}{Yongqi Fan}, \bibinfo{person}{Linjie Mu}, \bibinfo{person}{Ruoyu Liu}, \bibinfo{person}{Tong Ruan}, \bibinfo{person}{Shaoting Zhang}, {and} \bibinfo{person}{Xiaofan Zhang}.} \bibinfo{year}{2024}\natexlab{}.
\newblock \showarticletitle{Tool Calling: Enhancing Medication Consultation via Retrieval-Augmented Large Language Models}.
\newblock \bibinfo{journal}{\emph{arXiv preprint arXiv:2404.17897}} (\bibinfo{year}{2024}).
\newblock


\bibitem[Jian et~al\mbox{.}(2024)]%
        {jian2024tri}
\bibfield{author}{\bibinfo{person}{Chengtao Jian}, \bibinfo{person}{Kai Yang}, {and} \bibinfo{person}{Yang Jiao}.} \bibinfo{year}{2024}\natexlab{}.
\newblock \showarticletitle{Tri-Level Navigator: LLM-Empowered Tri-Level Learning for Time Series OOD Generalization}.
\newblock \bibinfo{journal}{\emph{Advances in Neural Information Processing Systems}}  \bibinfo{volume}{37} (\bibinfo{year}{2024}), \bibinfo{pages}{110613--110642}.
\newblock


\bibitem[Jiao et~al\mbox{.}(2022a)]%
        {jiao2022timeautoad}
\bibfield{author}{\bibinfo{person}{Yang Jiao}, \bibinfo{person}{Kai Yang}, \bibinfo{person}{Dongjing Song}, {and} \bibinfo{person}{Dacheng Tao}.} \bibinfo{year}{2022}\natexlab{a}.
\newblock \showarticletitle{Timeautoad: Autonomous anomaly detection with self-supervised contrastive loss for multivariate time series}.
\newblock \bibinfo{journal}{\emph{IEEE Transactions on Network Science and Engineering}} \bibinfo{volume}{9}, \bibinfo{number}{3} (\bibinfo{year}{2022}), \bibinfo{pages}{1604--1619}.
\newblock


\bibitem[Jiao et~al\mbox{.}({[n.\,d.]})]%
        {jiaoasynchronous}
\bibfield{author}{\bibinfo{person}{Yang Jiao}, \bibinfo{person}{Kai Yang}, \bibinfo{person}{Tiancheng Wu}, \bibinfo{person}{Dongjin Song}, {and} \bibinfo{person}{Chengtao Jian}.} \bibinfo{year}{[n.\,d.]}\natexlab{}.
\newblock \showarticletitle{Asynchronous Distributed Bilevel Optimization}. In \bibinfo{booktitle}{\emph{The Eleventh International Conference on Learning Representations}}.
\newblock


\bibitem[Jiao et~al\mbox{.}(2022b)]%
        {jiao2022asynchronous}
\bibfield{author}{\bibinfo{person}{Yang Jiao}, \bibinfo{person}{Kai Yang}, \bibinfo{person}{Tiancheng Wu}, \bibinfo{person}{Dongjin Song}, {and} \bibinfo{person}{Chengtao Jian}.} \bibinfo{year}{2022}\natexlab{b}.
\newblock \showarticletitle{Asynchronous Distributed Bilevel Optimization}. In \bibinfo{booktitle}{\emph{The Eleventh International Conference on Learning Representations}}.
\newblock


\bibitem[Khatoon et~al\mbox{.}(2022)]%
        {khatoon2022social}
\bibfield{author}{\bibinfo{person}{Shaheen Khatoon}, \bibinfo{person}{Amna Asif}, \bibinfo{person}{Md~Maruf Hasan}, {and} \bibinfo{person}{Majed Alshamari}.} \bibinfo{year}{2022}\natexlab{}.
\newblock \showarticletitle{Social media-based intelligence for disaster response and management in smart cities}.
\newblock In \bibinfo{booktitle}{\emph{Artificial Intelligence, Machine Learning, and Optimization Tools for Smart Cities: Designing for Sustainability}}. \bibinfo{publisher}{Springer}, \bibinfo{pages}{211--235}.
\newblock


\bibitem[Kim and Hospedales(2025)]%
        {kim2025stochastic}
\bibfield{author}{\bibinfo{person}{Minyoung Kim} {and} \bibinfo{person}{Timothy Hospedales}.} \bibinfo{year}{2025}\natexlab{}.
\newblock \showarticletitle{A Stochastic Approach to Bi-Level Optimization for Hyperparameter Optimization and Meta Learning}. In \bibinfo{booktitle}{\emph{Proceedings of the AAAI Conference on Artificial Intelligence}}, Vol.~\bibinfo{volume}{39}. \bibinfo{pages}{17913--17920}.
\newblock


\bibitem[Kourtit(2021)]%
        {kourtit2021city}
\bibfield{author}{\bibinfo{person}{Karima Kourtit}.} \bibinfo{year}{2021}\natexlab{}.
\newblock \showarticletitle{City intelligence for enhancing urban performance value: a conceptual study on data decomposition in smart cities}.
\newblock \bibinfo{journal}{\emph{Asia-Pacific Journal of Regional Science}} \bibinfo{volume}{5}, \bibinfo{number}{1} (\bibinfo{year}{2021}), \bibinfo{pages}{191--222}.
\newblock


\bibitem[Latif et~al\mbox{.}(2023)]%
        {latif2023artificial}
\bibfield{author}{\bibinfo{person}{Ehsan Latif}, \bibinfo{person}{Gengchen Mai}, \bibinfo{person}{Matthew Nyaaba}, \bibinfo{person}{Xuansheng Wu}, \bibinfo{person}{Ninghao Liu}, \bibinfo{person}{Guoyu Lu}, \bibinfo{person}{Sheng Li}, \bibinfo{person}{Tianming Liu}, {and} \bibinfo{person}{Xiaoming Zhai}.} \bibinfo{year}{2023}\natexlab{}.
\newblock \showarticletitle{Artificial general intelligence (AGI) for education}.
\newblock \bibinfo{journal}{\emph{arXiv preprint arXiv:2304.12479}}  \bibinfo{volume}{1} (\bibinfo{year}{2023}).
\newblock


\bibitem[Lee et~al\mbox{.}(2025)]%
        {lee2025rag}
\bibfield{author}{\bibinfo{person}{Namkyeong Lee}, \bibinfo{person}{Edward De~Brouwer}, \bibinfo{person}{Ehsan Hajiramezanali}, \bibinfo{person}{Tommaso Biancalani}, \bibinfo{person}{Chanyoung Park}, {and} \bibinfo{person}{Gabriele Scalia}.} \bibinfo{year}{2025}\natexlab{}.
\newblock \showarticletitle{RAG-Enhanced Collaborative LLM Agents for Drug Discovery}.
\newblock \bibinfo{journal}{\emph{arXiv preprint arXiv:2502.17506}} (\bibinfo{year}{2025}).
\newblock


\bibitem[Lesort et~al\mbox{.}(2020)]%
        {lesort2020continual}
\bibfield{author}{\bibinfo{person}{Timoth{\'e}e Lesort}, \bibinfo{person}{Vincenzo Lomonaco}, \bibinfo{person}{Andrei Stoian}, \bibinfo{person}{Davide Maltoni}, \bibinfo{person}{David Filliat}, {and} \bibinfo{person}{Natalia D{\'\i}az-Rodr{\'\i}guez}.} \bibinfo{year}{2020}\natexlab{}.
\newblock \showarticletitle{Continual learning for robotics: Definition, framework, learning strategies, opportunities and challenges}.
\newblock \bibinfo{journal}{\emph{Information fusion}}  \bibinfo{volume}{58} (\bibinfo{year}{2020}), \bibinfo{pages}{52--68}.
\newblock


\bibitem[Lewis et~al\mbox{.}(2020)]%
        {lewis2020retrieval}
\bibfield{author}{\bibinfo{person}{Patrick Lewis}, \bibinfo{person}{Ethan Perez}, \bibinfo{person}{Aleksandra Piktus}, \bibinfo{person}{Fabio Petroni}, \bibinfo{person}{Vladimir Karpukhin}, \bibinfo{person}{Naman Goyal}, \bibinfo{person}{Heinrich K{\"u}ttler}, \bibinfo{person}{Mike Lewis}, \bibinfo{person}{Wen-tau Yih}, \bibinfo{person}{Tim Rockt{\"a}schel}, {et~al\mbox{.}}} \bibinfo{year}{2020}\natexlab{}.
\newblock \showarticletitle{Retrieval-augmented generation for knowledge-intensive nlp tasks}.
\newblock \bibinfo{journal}{\emph{Advances in neural information processing systems}}  \bibinfo{volume}{33} (\bibinfo{year}{2020}), \bibinfo{pages}{9459--9474}.
\newblock


\bibitem[Li(2025)]%
        {li2025review}
\bibfield{author}{\bibinfo{person}{Xinzhe Li}.} \bibinfo{year}{2025}\natexlab{}.
\newblock \showarticletitle{A Review of Prominent Paradigms for LLM-Based Agents: Tool Use, Planning (Including RAG), and Feedback Learning}. In \bibinfo{booktitle}{\emph{Proceedings of the 31st International Conference on Computational Linguistics}}. \bibinfo{pages}{9760--9779}.
\newblock


\bibitem[Li et~al\mbox{.}(2025)]%
        {li2025survey}
\bibfield{author}{\bibinfo{person}{Xiaopeng Li}, \bibinfo{person}{Pengyue Jia}, \bibinfo{person}{Derong Xu}, \bibinfo{person}{Yi Wen}, \bibinfo{person}{Yingyi Zhang}, \bibinfo{person}{Wenlin Zhang}, \bibinfo{person}{Wanyu Wang}, \bibinfo{person}{Yichao Wang}, \bibinfo{person}{Zhaocheng Du}, \bibinfo{person}{Xiangyang Li}, {et~al\mbox{.}}} \bibinfo{year}{2025}\natexlab{}.
\newblock \showarticletitle{A Survey of Personalization: From RAG to Agent}.
\newblock \bibinfo{journal}{\emph{arXiv preprint arXiv:2504.10147}} (\bibinfo{year}{2025}).
\newblock


\bibitem[Li et~al\mbox{.}(2024)]%
        {li2024artificial}
\bibfield{author}{\bibinfo{person}{Xiang Li}, \bibinfo{person}{Lin Zhao}, \bibinfo{person}{Lu Zhang}, \bibinfo{person}{Zihao Wu}, \bibinfo{person}{Zhengliang Liu}, \bibinfo{person}{Hanqi Jiang}, \bibinfo{person}{Chao Cao}, \bibinfo{person}{Shaochen Xu}, \bibinfo{person}{Yiwei Li}, \bibinfo{person}{Haixing Dai}, {et~al\mbox{.}}} \bibinfo{year}{2024}\natexlab{}.
\newblock \showarticletitle{Artificial general intelligence for medical imaging analysis}.
\newblock \bibinfo{journal}{\emph{IEEE Reviews in Biomedical Engineering}} (\bibinfo{year}{2024}).
\newblock


\bibitem[Liu et~al\mbox{.}(2019)]%
        {liu2019seasonal}
\bibfield{author}{\bibinfo{person}{Huimin Liu}, \bibinfo{person}{Qingming Zhan}, \bibinfo{person}{Sihang Gao}, {and} \bibinfo{person}{Chen Yang}.} \bibinfo{year}{2019}\natexlab{}.
\newblock \showarticletitle{Seasonal variation of the spatially non-stationary association between land surface temperature and urban landscape}.
\newblock \bibinfo{journal}{\emph{Remote Sensing}} \bibinfo{volume}{11}, \bibinfo{number}{9} (\bibinfo{year}{2019}), \bibinfo{pages}{1016}.
\newblock


\bibitem[Liu et~al\mbox{.}(2021)]%
        {liu2021investigating}
\bibfield{author}{\bibinfo{person}{Risheng Liu}, \bibinfo{person}{Jiaxin Gao}, \bibinfo{person}{Jin Zhang}, \bibinfo{person}{Deyu Meng}, {and} \bibinfo{person}{Zhouchen Lin}.} \bibinfo{year}{2021}\natexlab{}.
\newblock \showarticletitle{Investigating bi-level optimization for learning and vision from a unified perspective: A survey and beyond}.
\newblock \bibinfo{journal}{\emph{IEEE Transactions on Pattern Analysis and Machine Intelligence}} \bibinfo{volume}{44}, \bibinfo{number}{12} (\bibinfo{year}{2021}), \bibinfo{pages}{10045--10067}.
\newblock


\bibitem[Liu et~al\mbox{.}(2023)]%
        {liu2023knowsite}
\bibfield{author}{\bibinfo{person}{Yu Liu}, \bibinfo{person}{Jingtao Ding}, {and} \bibinfo{person}{Yong Li}.} \bibinfo{year}{2023}\natexlab{}.
\newblock \showarticletitle{Knowsite: Leveraging urban knowledge graph for site selection}. In \bibinfo{booktitle}{\emph{Proceedings of the 31st ACM International Conference on Advances in Geographic Information Systems}}. \bibinfo{pages}{1--12}.
\newblock


\bibitem[Longley and Tob{\'o}n(2004)]%
        {longley2004spatial}
\bibfield{author}{\bibinfo{person}{Paul~A Longley} {and} \bibinfo{person}{Carolina Tob{\'o}n}.} \bibinfo{year}{2004}\natexlab{}.
\newblock \showarticletitle{Spatial dependence and heterogeneity in patterns of hardship: an intra-urban analysis}.
\newblock \bibinfo{journal}{\emph{Annals of the Association of American Geographers}} \bibinfo{volume}{94}, \bibinfo{number}{3} (\bibinfo{year}{2004}), \bibinfo{pages}{503--519}.
\newblock


\bibitem[Ma and Yang(2023)]%
        {ma2023metastnet}
\bibfield{author}{\bibinfo{person}{Hui Ma} {and} \bibinfo{person}{Kai Yang}.} \bibinfo{year}{2023}\natexlab{}.
\newblock \showarticletitle{Metastnet: Multimodal meta-learning for cellular traffic conformal prediction}.
\newblock \bibinfo{journal}{\emph{IEEE Transactions on Network Science and Engineering}} \bibinfo{volume}{11}, \bibinfo{number}{2} (\bibinfo{year}{2023}), \bibinfo{pages}{1999--2011}.
\newblock


\bibitem[Ma et~al\mbox{.}(2023)]%
        {ma2023cellular}
\bibfield{author}{\bibinfo{person}{Hui Ma}, \bibinfo{person}{Kai Yang}, {and} \bibinfo{person}{Man-On Pun}.} \bibinfo{year}{2023}\natexlab{}.
\newblock \showarticletitle{Cellular traffic prediction via deep state space models with attention mechanism}.
\newblock \bibinfo{journal}{\emph{Computer Communications}}  \bibinfo{volume}{197} (\bibinfo{year}{2023}), \bibinfo{pages}{276--283}.
\newblock


\bibitem[Mahor et~al\mbox{.}(2023)]%
        {mahor2023iot}
\bibfield{author}{\bibinfo{person}{Vinod Mahor}, \bibinfo{person}{Romil Rawat}, \bibinfo{person}{Anil Kumar}, \bibinfo{person}{Bhagwati Garg}, \bibinfo{person}{Kiran Pachlasiya}, {et~al\mbox{.}}} \bibinfo{year}{2023}\natexlab{}.
\newblock \showarticletitle{IoT and artificial intelligence techniques for public safety and security}.
\newblock In \bibinfo{booktitle}{\emph{Smart urban computing applications}}. \bibinfo{publisher}{River Publishers}, \bibinfo{pages}{111--126}.
\newblock


\bibitem[Men{\'e}ndez et~al\mbox{.}(1997)]%
        {menendez1997jensen}
\bibfield{author}{\bibinfo{person}{Mar{\'\i}a~Luisa Men{\'e}ndez}, \bibinfo{person}{Julio~Angel Pardo}, \bibinfo{person}{Leandro Pardo}, {and} \bibinfo{person}{Mar{\'\i}a del~C Pardo}.} \bibinfo{year}{1997}\natexlab{}.
\newblock \showarticletitle{The jensen-shannon divergence}.
\newblock \bibinfo{journal}{\emph{Journal of the Franklin Institute}} \bibinfo{volume}{334}, \bibinfo{number}{2} (\bibinfo{year}{1997}), \bibinfo{pages}{307--318}.
\newblock


\bibitem[Morris et~al\mbox{.}(2024)]%
        {morris2024position}
\bibfield{author}{\bibinfo{person}{Meredith~Ringel Morris}, \bibinfo{person}{Jascha Sohl-Dickstein}, \bibinfo{person}{Noah Fiedel}, \bibinfo{person}{Tris Warkentin}, \bibinfo{person}{Allan Dafoe}, \bibinfo{person}{Aleksandra Faust}, \bibinfo{person}{Clement Farabet}, {and} \bibinfo{person}{Shane Legg}.} \bibinfo{year}{2024}\natexlab{}.
\newblock \showarticletitle{Position: Levels of AGI for operationalizing progress on the path to AGI}. In \bibinfo{booktitle}{\emph{Forty-first International Conference on Machine Learning}}.
\newblock


\bibitem[Nama et~al\mbox{.}(2021)]%
        {nama2021machine}
\bibfield{author}{\bibinfo{person}{Mahima Nama}, \bibinfo{person}{Ankita Nath}, \bibinfo{person}{Nancy Bechra}, \bibinfo{person}{Jitendra Bhatia}, \bibinfo{person}{Sudeep Tanwar}, \bibinfo{person}{Manish Chaturvedi}, {and} \bibinfo{person}{Balqies Sadoun}.} \bibinfo{year}{2021}\natexlab{}.
\newblock \showarticletitle{Machine learning-based traffic scheduling techniques for intelligent transportation system: Opportunities and challenges}.
\newblock \bibinfo{journal}{\emph{International Journal of Communication Systems}} \bibinfo{volume}{34}, \bibinfo{number}{9} (\bibinfo{year}{2021}), \bibinfo{pages}{e4814}.
\newblock


\bibitem[Nguyen et~al\mbox{.}(2019)]%
        {nguyen2019toward}
\bibfield{author}{\bibinfo{person}{Cuong~V Nguyen}, \bibinfo{person}{Alessandro Achille}, \bibinfo{person}{Michael Lam}, \bibinfo{person}{Tal Hassner}, \bibinfo{person}{Vijay Mahadevan}, {and} \bibinfo{person}{Stefano Soatto}.} \bibinfo{year}{2019}\natexlab{}.
\newblock \showarticletitle{Toward understanding catastrophic forgetting in continual learning}.
\newblock \bibinfo{journal}{\emph{arXiv preprint arXiv:1908.01091}} (\bibinfo{year}{2019}).
\newblock


\bibitem[Nisha et~al\mbox{.}(2022)]%
        {nisha2022bilevel}
\bibfield{author}{\bibinfo{person}{Tarannum Nisha}, \bibinfo{person}{Duong~Tung Nguyen}, {and} \bibinfo{person}{Vijay~K Bhargava}.} \bibinfo{year}{2022}\natexlab{}.
\newblock \showarticletitle{A bilevel programming framework for joint edge resource management and pricing}.
\newblock \bibinfo{journal}{\emph{IEEE Internet of Things Journal}} \bibinfo{volume}{9}, \bibinfo{number}{18} (\bibinfo{year}{2022}), \bibinfo{pages}{17280--17291}.
\newblock


\bibitem[Pacheco~Rocha et~al\mbox{.}(2022)]%
        {pacheco2022systematic}
\bibfield{author}{\bibinfo{person}{Nelson Pacheco~Rocha}, \bibinfo{person}{Ana Dias}, \bibinfo{person}{Gon{\c{c}}alo Santinha}, \bibinfo{person}{M{\'a}rio Rodrigues}, \bibinfo{person}{Carlos Rodrigues}, \bibinfo{person}{Alexandra Queir{\'o}s}, \bibinfo{person}{Rute Bastardo}, {and} \bibinfo{person}{Jo{\~a}o Pav{\~a}o}.} \bibinfo{year}{2022}\natexlab{}.
\newblock \showarticletitle{Systematic literature review of context-awareness applications supported by smart cities’ infrastructures}.
\newblock \bibinfo{journal}{\emph{SN Applied Sciences}} \bibinfo{volume}{4}, \bibinfo{number}{4} (\bibinfo{year}{2022}), \bibinfo{pages}{90}.
\newblock


\bibitem[Patil et~al\mbox{.}(2024)]%
        {patil2024gorilla}
\bibfield{author}{\bibinfo{person}{Shishir~G Patil}, \bibinfo{person}{Tianjun Zhang}, \bibinfo{person}{Xin Wang}, {and} \bibinfo{person}{Joseph~E Gonzalez}.} \bibinfo{year}{2024}\natexlab{}.
\newblock \showarticletitle{Gorilla: Large language model connected with massive apis}.
\newblock \bibinfo{journal}{\emph{Advances in Neural Information Processing Systems}}  \bibinfo{volume}{37} (\bibinfo{year}{2024}), \bibinfo{pages}{126544--126565}.
\newblock


\bibitem[Pomponi et~al\mbox{.}(2020)]%
        {pomponi2020efficient}
\bibfield{author}{\bibinfo{person}{Jary Pomponi}, \bibinfo{person}{Simone Scardapane}, \bibinfo{person}{Vincenzo Lomonaco}, {and} \bibinfo{person}{Aurelio Uncini}.} \bibinfo{year}{2020}\natexlab{}.
\newblock \showarticletitle{Efficient continual learning in neural networks with embedding regularization}.
\newblock \bibinfo{journal}{\emph{Neurocomputing}}  \bibinfo{volume}{397} (\bibinfo{year}{2020}), \bibinfo{pages}{139--148}.
\newblock


\bibitem[Poredi et~al\mbox{.}(2023)]%
        {poredi2023enhance}
\bibfield{author}{\bibinfo{person}{Nihal Poredi}, \bibinfo{person}{Yu Chen}, \bibinfo{person}{Xiaohua Li}, {and} \bibinfo{person}{Erik Blasch}.} \bibinfo{year}{2023}\natexlab{}.
\newblock \showarticletitle{Enhance public safety surveillance in smart cities by fusing optical and thermal cameras}. In \bibinfo{booktitle}{\emph{2023 26th International Conference on Information Fusion (FUSION)}}. IEEE, \bibinfo{pages}{1--7}.
\newblock


\bibitem[Qian et~al\mbox{.}(2019)]%
        {qian2019robust}
\bibfield{author}{\bibinfo{person}{Qi Qian}, \bibinfo{person}{Shenghuo Zhu}, \bibinfo{person}{Jiasheng Tang}, \bibinfo{person}{Rong Jin}, \bibinfo{person}{Baigui Sun}, {and} \bibinfo{person}{Hao Li}.} \bibinfo{year}{2019}\natexlab{}.
\newblock \showarticletitle{Robust optimization over multiple domains}. In \bibinfo{booktitle}{\emph{Proceedings of the AAAI Conference on Artificial Intelligence}}, Vol.~\bibinfo{volume}{33}. \bibinfo{pages}{4739--4746}.
\newblock


\bibitem[Qu et~al\mbox{.}(2025)]%
        {qu2025recent}
\bibfield{author}{\bibinfo{person}{Haoxuan Qu}, \bibinfo{person}{Hossein Rahmani}, \bibinfo{person}{Li Xu}, \bibinfo{person}{Bryan Williams}, {and} \bibinfo{person}{Jun Liu}.} \bibinfo{year}{2025}\natexlab{}.
\newblock \showarticletitle{Recent advances of continual learning in computer vision: An overview}.
\newblock \bibinfo{journal}{\emph{IET Computer Vision}} \bibinfo{volume}{19}, \bibinfo{number}{1} (\bibinfo{year}{2025}), \bibinfo{pages}{e70013}.
\newblock


\bibitem[Ramana et~al\mbox{.}(2023)]%
        {ramana2023vision}
\bibfield{author}{\bibinfo{person}{Kadiyala Ramana}, \bibinfo{person}{Gautam Srivastava}, \bibinfo{person}{Madapuri~Rudra Kumar}, \bibinfo{person}{Thippa~Reddy Gadekallu}, \bibinfo{person}{Jerry Chun-Wei Lin}, \bibinfo{person}{Mamoun Alazab}, {and} \bibinfo{person}{Celestine Iwendi}.} \bibinfo{year}{2023}\natexlab{}.
\newblock \showarticletitle{A vision transformer approach for traffic congestion prediction in urban areas}.
\newblock \bibinfo{journal}{\emph{IEEE Transactions on Intelligent Transportation Systems}} \bibinfo{volume}{24}, \bibinfo{number}{4} (\bibinfo{year}{2023}), \bibinfo{pages}{3922--3934}.
\newblock


\bibitem[Ravish and Swamy(2021)]%
        {ravish2021intelligent}
\bibfield{author}{\bibinfo{person}{Roopa Ravish} {and} \bibinfo{person}{Shanta~Ranga Swamy}.} \bibinfo{year}{2021}\natexlab{}.
\newblock \showarticletitle{Intelligent traffic management: A review of challenges, solutions, and future perspectives}.
\newblock \bibinfo{journal}{\emph{Transport and Telecommunication}} \bibinfo{volume}{22}, \bibinfo{number}{2} (\bibinfo{year}{2021}), \bibinfo{pages}{163--182}.
\newblock


\bibitem[Roychowdhury et~al\mbox{.}(2024)]%
        {roychowdhury2024evaluation}
\bibfield{author}{\bibinfo{person}{Sujoy Roychowdhury}, \bibinfo{person}{Sumit Soman}, \bibinfo{person}{HG Ranjani}, \bibinfo{person}{Neeraj Gunda}, \bibinfo{person}{Vansh Chhabra}, {and} \bibinfo{person}{Sai~Krishna Bala}.} \bibinfo{year}{2024}\natexlab{}.
\newblock \showarticletitle{Evaluation of rag metrics for question answering in the telecom domain}.
\newblock \bibinfo{journal}{\emph{arXiv preprint arXiv:2407.12873}} (\bibinfo{year}{2024}).
\newblock


\bibitem[Rui et~al\mbox{.}(2023)]%
        {rui2023dilrs}
\bibfield{author}{\bibinfo{person}{Xue Rui}, \bibinfo{person}{Ziqiang Li}, \bibinfo{person}{Yang Cao}, \bibinfo{person}{Ziyang Li}, {and} \bibinfo{person}{Weiguo Song}.} \bibinfo{year}{2023}\natexlab{}.
\newblock \showarticletitle{DILRS: Domain-incremental learning for semantic segmentation in multi-source remote sensing data}.
\newblock \bibinfo{journal}{\emph{Remote Sensing}} \bibinfo{volume}{15}, \bibinfo{number}{10} (\bibinfo{year}{2023}), \bibinfo{pages}{2541}.
\newblock


\bibitem[R{\"u}schendorf(1985)]%
        {ruschendorf1985wasserstein}
\bibfield{author}{\bibinfo{person}{Ludger R{\"u}schendorf}.} \bibinfo{year}{1985}\natexlab{}.
\newblock \showarticletitle{The Wasserstein distance and approximation theorems}.
\newblock \bibinfo{journal}{\emph{Probability Theory and Related Fields}} \bibinfo{volume}{70}, \bibinfo{number}{1} (\bibinfo{year}{1985}), \bibinfo{pages}{117--129}.
\newblock


\bibitem[Srivastava et~al\mbox{.}(2017)]%
        {srivastava2017safety}
\bibfield{author}{\bibinfo{person}{Shweta Srivastava}, \bibinfo{person}{Aditya Bisht}, {and} \bibinfo{person}{Neetu Narayan}.} \bibinfo{year}{2017}\natexlab{}.
\newblock \showarticletitle{Safety and security in smart cities using artificial intelligence—A review}. In \bibinfo{booktitle}{\emph{2017 7th international conference on cloud computing, data science \& engineering-confluence}}. IEEE, \bibinfo{pages}{130--133}.
\newblock


\bibitem[Touvron et~al\mbox{.}(2023)]%
        {touvron2023llama}
\bibfield{author}{\bibinfo{person}{Hugo Touvron}, \bibinfo{person}{Louis Martin}, \bibinfo{person}{Kevin Stone}, \bibinfo{person}{Peter Albert}, \bibinfo{person}{Amjad Almahairi}, \bibinfo{person}{Yasmine Babaei}, \bibinfo{person}{Nikolay Bashlykov}, \bibinfo{person}{Soumya Batra}, \bibinfo{person}{Prajjwal Bhargava}, \bibinfo{person}{Shruti Bhosale}, {et~al\mbox{.}}} \bibinfo{year}{2023}\natexlab{}.
\newblock \showarticletitle{Llama 2: Open foundation and fine-tuned chat models}.
\newblock \bibinfo{journal}{\emph{arXiv preprint arXiv:2307.09288}} (\bibinfo{year}{2023}).
\newblock


\bibitem[Wang et~al\mbox{.}(2024a)]%
        {wang2024comprehensive}
\bibfield{author}{\bibinfo{person}{Liyuan Wang}, \bibinfo{person}{Xingxing Zhang}, \bibinfo{person}{Hang Su}, {and} \bibinfo{person}{Jun Zhu}.} \bibinfo{year}{2024}\natexlab{a}.
\newblock \showarticletitle{A comprehensive survey of continual learning: Theory, method and application}.
\newblock \bibinfo{journal}{\emph{IEEE Transactions on Pattern Analysis and Machine Intelligence}} (\bibinfo{year}{2024}).
\newblock


\bibitem[Wang et~al\mbox{.}(2022)]%
        {wang2022memory}
\bibfield{author}{\bibinfo{person}{Liyuan Wang}, \bibinfo{person}{Xingxing Zhang}, \bibinfo{person}{Kuo Yang}, \bibinfo{person}{Longhui Yu}, \bibinfo{person}{Chongxuan Li}, \bibinfo{person}{Lanqing Hong}, \bibinfo{person}{Shifeng Zhang}, \bibinfo{person}{Zhenguo Li}, \bibinfo{person}{Yi Zhong}, {and} \bibinfo{person}{Jun Zhu}.} \bibinfo{year}{2022}\natexlab{}.
\newblock \showarticletitle{Memory replay with data compression for continual learning}.
\newblock \bibinfo{journal}{\emph{arXiv preprint arXiv:2202.06592}} (\bibinfo{year}{2022}).
\newblock


\bibitem[Wang et~al\mbox{.}(2024b)]%
        {wang2024reinforcement}
\bibfield{author}{\bibinfo{person}{Shuhe Wang}, \bibinfo{person}{Shengyu Zhang}, \bibinfo{person}{Jie Zhang}, \bibinfo{person}{Runyi Hu}, \bibinfo{person}{Xiaoya Li}, \bibinfo{person}{Tianwei Zhang}, \bibinfo{person}{Jiwei Li}, \bibinfo{person}{Fei Wu}, \bibinfo{person}{Guoyin Wang}, {and} \bibinfo{person}{Eduard Hovy}.} \bibinfo{year}{2024}\natexlab{b}.
\newblock \showarticletitle{Reinforcement learning enhanced llms: A survey}.
\newblock \bibinfo{journal}{\emph{arXiv preprint arXiv:2412.10400}} (\bibinfo{year}{2024}).
\newblock


\bibitem[Wu et~al\mbox{.}(2024)]%
        {wu2024avatar}
\bibfield{author}{\bibinfo{person}{Shirley Wu}, \bibinfo{person}{Shiyu Zhao}, \bibinfo{person}{Qian Huang}, \bibinfo{person}{Kexin Huang}, \bibinfo{person}{Michihiro Yasunaga}, \bibinfo{person}{Kaidi Cao}, \bibinfo{person}{Vassilis Ioannidis}, \bibinfo{person}{Karthik Subbian}, \bibinfo{person}{Jure Leskovec}, {and} \bibinfo{person}{James~Y Zou}.} \bibinfo{year}{2024}\natexlab{}.
\newblock \showarticletitle{Avatar: Optimizing llm agents for tool usage via contrastive reasoning}.
\newblock \bibinfo{journal}{\emph{Advances in Neural Information Processing Systems}}  \bibinfo{volume}{37} (\bibinfo{year}{2024}), \bibinfo{pages}{25981--26010}.
\newblock


\bibitem[Xia et~al\mbox{.}(2024)]%
        {xia2024rule}
\bibfield{author}{\bibinfo{person}{Peng Xia}, \bibinfo{person}{Kangyu Zhu}, \bibinfo{person}{Haoran Li}, \bibinfo{person}{Hongtu Zhu}, \bibinfo{person}{Yun Li}, \bibinfo{person}{Gang Li}, \bibinfo{person}{Linjun Zhang}, {and} \bibinfo{person}{Huaxiu Yao}.} \bibinfo{year}{2024}\natexlab{}.
\newblock \showarticletitle{Rule: Reliable multimodal rag for factuality in medical vision language models}. In \bibinfo{booktitle}{\emph{Proceedings of the 2024 Conference on Empirical Methods in Natural Language Processing}}. \bibinfo{pages}{1081--1093}.
\newblock


\bibitem[Xu et~al\mbox{.}(2023)]%
        {xu2023urban}
\bibfield{author}{\bibinfo{person}{Fengli Xu}, \bibinfo{person}{Jun Zhang}, \bibinfo{person}{Chen Gao}, \bibinfo{person}{Jie Feng}, {and} \bibinfo{person}{Yong Li}.} \bibinfo{year}{2023}\natexlab{}.
\newblock \showarticletitle{Urban Generative Intelligence (UGI): A Foundational Platform for Agents in Embodied City Environment}.
\newblock \bibinfo{journal}{\emph{CoRR}} (\bibinfo{year}{2023}).
\newblock


\bibitem[Xu et~al\mbox{.}(2024)]%
        {xu2024retrieval}
\bibfield{author}{\bibinfo{person}{Zhentao Xu}, \bibinfo{person}{Mark~Jerome Cruz}, \bibinfo{person}{Matthew Guevara}, \bibinfo{person}{Tie Wang}, \bibinfo{person}{Manasi Deshpande}, \bibinfo{person}{Xiaofeng Wang}, {and} \bibinfo{person}{Zheng Li}.} \bibinfo{year}{2024}\natexlab{}.
\newblock \showarticletitle{Retrieval-augmented generation with knowledge graphs for customer service question answering}. In \bibinfo{booktitle}{\emph{Proceedings of the 47th International ACM SIGIR Conference on Research and Development in Information Retrieval}}. \bibinfo{pages}{2905--2909}.
\newblock


\bibitem[Yang et~al\mbox{.}(2022)]%
        {yang2022survey}
\bibfield{author}{\bibinfo{person}{Fei Yang}, \bibinfo{person}{Yixin Hua}, \bibinfo{person}{Xiang Li}, \bibinfo{person}{Zhenkai Yang}, \bibinfo{person}{Xinkai Yu}, {and} \bibinfo{person}{Teng Fei}.} \bibinfo{year}{2022}\natexlab{}.
\newblock \showarticletitle{A survey on multisource heterogeneous urban sensor access and data management technologies}.
\newblock \bibinfo{journal}{\emph{Measurement: Sensors}}  \bibinfo{volume}{19} (\bibinfo{year}{2022}), \bibinfo{pages}{100061}.
\newblock


\bibitem[Yang et~al\mbox{.}(2014)]%
        {yang2014distributed}
\bibfield{author}{\bibinfo{person}{Kai Yang}, \bibinfo{person}{Jianwei Huang}, \bibinfo{person}{Yihong Wu}, \bibinfo{person}{Xiaodong Wang}, {and} \bibinfo{person}{Mung Chiang}.} \bibinfo{year}{2014}\natexlab{}.
\newblock \showarticletitle{Distributed robust optimization (DRO), part I: Framework and example}.
\newblock \bibinfo{journal}{\emph{Optimization and Engineering}}  \bibinfo{volume}{15} (\bibinfo{year}{2014}), \bibinfo{pages}{35--67}.
\newblock


\bibitem[Yang et~al\mbox{.}(2008)]%
        {yang2008distributed}
\bibfield{author}{\bibinfo{person}{Kai Yang}, \bibinfo{person}{Yihong Wu}, \bibinfo{person}{Jianwei Huang}, \bibinfo{person}{Xiaodong Wang}, {and} \bibinfo{person}{Sergio Verd{\'u}}.} \bibinfo{year}{2008}\natexlab{}.
\newblock \showarticletitle{Distributed robust optimization for communication networks}. In \bibinfo{booktitle}{\emph{IEEE INFOCOM 2008-The 27th Conference on Computer Communications}}. IEEE, \bibinfo{pages}{1157--1165}.
\newblock


\bibitem[Yang et~al\mbox{.}(2024)]%
        {yang2024continual}
\bibfield{author}{\bibinfo{person}{Li Yang}, \bibinfo{person}{Zhipeng Luo}, \bibinfo{person}{Shiming Zhang}, \bibinfo{person}{Fei Teng}, {and} \bibinfo{person}{Tianrui Li}.} \bibinfo{year}{2024}\natexlab{}.
\newblock \showarticletitle{Continual Learning for Smart City: A Survey}.
\newblock \bibinfo{journal}{\emph{IEEE Transactions on Knowledge and Data Engineering}} (\bibinfo{year}{2024}).
\newblock


\bibitem[Yang et~al\mbox{.}(2019)]%
        {yang2019mares}
\bibfield{author}{\bibinfo{person}{Min Yang}, \bibinfo{person}{Wenting Tu}, \bibinfo{person}{Qiang Qu}, \bibinfo{person}{Kai Lei}, \bibinfo{person}{Xiaojun Chen}, \bibinfo{person}{Jia Zhu}, {and} \bibinfo{person}{Ying Shen}.} \bibinfo{year}{2019}\natexlab{}.
\newblock \showarticletitle{MARES: multitask learning algorithm for Web-scale real-time event summarization}.
\newblock \bibinfo{journal}{\emph{World Wide Web}}  \bibinfo{volume}{22} (\bibinfo{year}{2019}), \bibinfo{pages}{499--515}.
\newblock


\bibitem[Zareapoor et~al\mbox{.}(2024)]%
        {zareapoor2024efficient}
\bibfield{author}{\bibinfo{person}{Masoumeh Zareapoor}, \bibinfo{person}{Pourya Shamsolmoali}, {and} \bibinfo{person}{Fateme Vesaghati}.} \bibinfo{year}{2024}\natexlab{}.
\newblock \showarticletitle{Efficient Routing in Sparse Mixture-of-Experts}. In \bibinfo{booktitle}{\emph{2024 International Joint Conference on Neural Networks (IJCNN)}}. IEEE, \bibinfo{pages}{1--8}.
\newblock


\bibitem[Zhang et~al\mbox{.}(2024c)]%
        {zhang2024cppo}
\bibfield{author}{\bibinfo{person}{Han Zhang}, \bibinfo{person}{Yu Lei}, \bibinfo{person}{Lin Gui}, \bibinfo{person}{Min Yang}, \bibinfo{person}{Yulan He}, \bibinfo{person}{Hui Wang}, {and} \bibinfo{person}{Ruifeng Xu}.} \bibinfo{year}{2024}\natexlab{c}.
\newblock \showarticletitle{Cppo: Continual learning for reinforcement learning with human feedback}. In \bibinfo{booktitle}{\emph{The Twelfth International Conference on Learning Representations}}.
\newblock


\bibitem[Zhang et~al\mbox{.}(2024a)]%
        {zhang2024multi}
\bibfield{author}{\bibinfo{person}{Rongyu Zhang}, \bibinfo{person}{Yun Chen}, \bibinfo{person}{Chenrui Wu}, \bibinfo{person}{Fangxin Wang}, {and} \bibinfo{person}{Bo Li}.} \bibinfo{year}{2024}\natexlab{a}.
\newblock \showarticletitle{Multi-level personalized federated learning on heterogeneous and long-tailed data}.
\newblock \bibinfo{journal}{\emph{IEEE Transactions on Mobile Computing}} \bibinfo{volume}{23}, \bibinfo{number}{12} (\bibinfo{year}{2024}), \bibinfo{pages}{12396--12409}.
\newblock


\bibitem[Zhang et~al\mbox{.}(2024b)]%
        {zhang2024urban}
\bibfield{author}{\bibinfo{person}{Weijia Zhang}, \bibinfo{person}{Jindong Han}, \bibinfo{person}{Zhao Xu}, \bibinfo{person}{Hang Ni}, \bibinfo{person}{Hao Liu}, {and} \bibinfo{person}{Hui Xiong}.} \bibinfo{year}{2024}\natexlab{b}.
\newblock \showarticletitle{Urban foundation models: A survey}. In \bibinfo{booktitle}{\emph{Proceedings of the 30th ACM SIGKDD Conference on Knowledge Discovery and Data Mining}}. \bibinfo{pages}{6633--6643}.
\newblock


\bibitem[Zhao et~al\mbox{.}(2024b)]%
        {zhao2024retrieval}
\bibfield{author}{\bibinfo{person}{Siyun Zhao}, \bibinfo{person}{Yuqing Yang}, \bibinfo{person}{Zilong Wang}, \bibinfo{person}{Zhiyuan He}, \bibinfo{person}{Luna~K Qiu}, {and} \bibinfo{person}{Lili Qiu}.} \bibinfo{year}{2024}\natexlab{b}.
\newblock \showarticletitle{Retrieval augmented generation (rag) and beyond: A comprehensive survey on how to make your llms use external data more wisely}.
\newblock \bibinfo{journal}{\emph{arXiv preprint arXiv:2409.14924}} (\bibinfo{year}{2024}).
\newblock


\bibitem[Zhao et~al\mbox{.}(2024a)]%
        {zhao2024let}
\bibfield{author}{\bibinfo{person}{Yuyue Zhao}, \bibinfo{person}{Jiancan Wu}, \bibinfo{person}{Xiang Wang}, \bibinfo{person}{Wei Tang}, \bibinfo{person}{Dingxian Wang}, {and} \bibinfo{person}{Maarten De~Rijke}.} \bibinfo{year}{2024}\natexlab{a}.
\newblock \showarticletitle{Let me do it for you: Towards llm empowered recommendation via tool learning}. In \bibinfo{booktitle}{\emph{Proceedings of the 47th International ACM SIGIR Conference on Research and Development in Information Retrieval}}. \bibinfo{pages}{1796--1806}.
\newblock


\end{thebibliography}

\end{document}